%% file: emnlp2023.tex
\pdfoutput=1

\documentclass[11pt]{article}

\usepackage[final]{EMNLP2023}

\usepackage{times}
\usepackage{latexsym}
\usepackage{amsmath}

\usepackage[T1]{fontenc}

\usepackage[utf8]{inputenc}

\usepackage{microtype}

\usepackage[inline,shortlabels]{enumitem}

\input{macros.tex}

\usepackage{graphicx}
\usepackage{url}
\usepackage{multirow}
\usepackage{booktabs}
\usepackage{longtable}

\usepackage{inconsolata}

\title{BioDEX: Large-Scale Biomedical Adverse Drug Event Extraction \\ for Real-World Pharmacovigilance}

\author{Karel D'Oosterlinck$^{1,*}$, Fran\c{c}ois Remy$^1$, Johannes Deleu$^1$, Thomas Demeester$^1$ \\  \textbf{Chris Develder$^1$, Klim Zaporojets$^2$, Aneiss Ghodsi$^3$, Simon Ellershaw$^3$, Jack Collins$^3$} \\ \textbf{Christopher Potts$^4$} \\ $^1$Ghent University -- imec\qquad$^2$Aarhus University\\$^3$Parexel AI Labs\qquad $^4$Stanford University \\
$^*$\texttt{karel.doosterlinck@ugent.be}
}

\begin{document}

\maketitle

\begin{abstract}

\input{content/abstract.tex}

\end{abstract}

\section{Introduction}

\input{content/introduction.tex}

\section{Pharmacovigilance Reporting} \label{sec:reporting}
\input{content/reporting}

\section{Related Work} \label{sec:related-work}
\input{content/related-work.tex}

\section{The \OurDataset{} Dataset} \label{sec:data}
\input{content/data.tex}

\section{Task and Metrics} \label{sec:metrics}
\input{content/metrics.tex}

\section{Experiments and Results} \label{sec:experiments}
\input{content/experiments.tex}

\section{Improving Pharmacovigilance} \label{sec:results}
\input{content/results.tex}

\section{Conclusion} \label{sec:conclusion}
\input{content/conclusion.tex}

\section{Limitations and Ethical Considerations} \label{sec:limitations}
\input{content/limitations.tex}

\section*{Acknowledgements}
\input{content/acknowledgements.tex}

\bibliography{anthology,custom}
\bibliographystyle{acl_natbib}

\newpage
\clearpage
\onecolumn
\appendix

\section{\OurDataset{} Report Schema}
\label{app:schema}
\input{appendix/schema}

\section{\OurDataset{} Article Schema}
\label{app:article-schema}
\input{appendix/article-schema}

\section{\OurDataset{} Report Attribute Frequencies}
\label{app:report-freq}
\input{appendix/report-freq}

\clearpage
\onecolumn
\section{Few-Shot Prompt} \label{app:prompt}
The prompt below is the one used for the in-context learning experiments. No effort was spent on prompt engineering and the demonstrations were randomly sampled from the training set. 
\\ \\
\input{tables/prompt}

\section{Example Outputs} \label{app:example-outputs}
The table below shows the first 10 examples of the \texttt{validation} split. The inputs are truncated after 2,500 characters. The associated simplified report is given as well as the model predictions for our best \texttt{FLAN-T5-Large} and \texttt{GPT-4} models.
\input{tables/prediction_comparison}
\clearpage

\section{Drug classification} \label{app:drug-classification}
\Figref{fig:drug-classification} describes the same experiment as in \secref{sec:results} but now performed on drugs instead of reactions.
\input{figures/drug-classification}

\end{document}

%% file: macros.tex
\def\OurDataset{BioDEX}
\def\PM{PubMed}

\def\tableref#1{Table~\ref{#1}}
\def\Tableref#1{Table~\ref{#1}}
\def\figref#1{Figure~\ref{#1}}
\def\Figref#1{Figure~\ref{#1}}

\def\secref#1{Section~\ref{#1}}

\def\appref#1{Appendix~\ref{#1}}
\def\Appref#1{Appendix~\ref{#1}}

\def\eqref#1{Eq.~\ref{#1}}

%% file: content/abstract.tex
Timely and accurate extraction of Adverse Drug Events (ADE) from biomedical literature is paramount for public safety, but involves slow and costly manual labor. We set out to improve drug safety monitoring (pharmacovigilance, PV) through the use of Natural Language Processing (NLP). We introduce \OurDataset{}, a large-scale resource for Biomedical adverse Drug Event eXtraction, rooted in the historical output of drug safety reporting in the U.S. \OurDataset{} consists of 65k abstracts and 19k full-text biomedical papers with 256k associated document-level safety reports created by medical experts. The core features of these reports include the reported weight, age, and biological sex of a patient, a set of drugs taken by the patient, the drug dosages, the reactions experienced, and whether the reaction was life threatening. In this work, we consider the task of predicting the core information of the report given its originating paper. We estimate human performance to be 72.0\% F1, whereas our best model achieves 59.1\% F1 (62.3 validation), indicating significant headroom. 
We also begin to explore ways in which these models could help professional PV reviewers.
Our code and data are available at \url{https://github.com/KarelDO/BioDEX}.

%% file: content/introduction.tex
In the United States, the Food and Drug Administration (FDA) mandates drug producers to monitor and report Adverse Drug Events (ADE) described in the biomedical literature. Such a report, called an Individual Case Safety Report (ICSR), is stored in the FDA Adverse Event Reporting System (FAERS; \citealt{fda_faers}), which is a cornerstone resource for drug safety research, also called pharmacovigilance (PV). 

\Figref{fig:ade-example} briefly summarizes the core information PV workers must extract from papers while constructing these reports. This includes a description of the patient in terms of reported weight, age, and biological sex, a list of drugs taken by the patient, and a list of adverse reactions experienced and whether they are considered serious.

\input{figures/ade-example.tex}
Drug manufacturers employ teams of experts to continually triage new papers and submit these reports. This is challenging work since it requires experts to survey entire biomedical papers and utilize their pre-existing knowledge about a drug of interest, its conventional indications, and its known adverse reactions.
Furthermore, manufacturers are placed under constant time pressure to keep up with the latest publications, since failure to report in a timely manner can lead to hefty fines and compromise public safety. This pressure has potential to increase in the near future: there has been a steady acceleration of biomedical research over the last few years (\figref{fig:article-report-trend}), and drug events are consistently under-reported \citep{alatawi2017empirical}.

\input{figures/article_report_trends}

In this work, we set out to improve the scalability and accuracy of PV using Natural Language Processing (NLP).
As a first step, we introduce \OurDataset{}, a large-scale dataset for document-level Biomedical adverse Drug Event eXtraction. \OurDataset{} consists of biomedical papers with associated expert-created drug safety reports. These reports were submitted to the FDA between 2012 and 2022 as part of real-world PV efforts. Thus, \OurDataset{} is grounded in the historical and regulatory context of drug safety monitoring in the U.S. \OurDataset{} contains \PM{} articles published between 1968 and 2022, with 65,648 articles having an abstract available and 19,433 featuring a full-text paper. In total, 256,240 reports are included (there can be multiple reports per article).

We evaluate the ability of language models (LMs) to fill out the core information of a report given a full-text article that is known to describe at least one ADE. We estimate a lower bound on human performance to be 72.0\%~F1. Our best model (a fine-tuned \texttt{FLAN-T5-Large}; \citealt{chung2022scaling}) attains 59.1\%~F1, indicating substantial additional room for improvement while also suggesting that models trained on \OurDataset{} are on a path to being useful tools for PV workers. Additionally, we evaluate the capability of OpenAI's GPT models (\texttt{text-davinci-002}, \texttt{text-davinci-003}, \texttt{gpt-3.5-turbo}, \texttt{gpt-4}; \citealt{brown2020language}) but find that they severely struggle with this task, attaining at most 53.1\%~F1. 

Our models can aid drug safety research efforts today. An important use-case for drug safety research is efficiently finding papers that describe an adverse event with regard to a specific drug or reaction. Conventional search baselines suffer from low precision, since mentioned drugs and reactions are only rarely involved in an adverse event. Our models are specifically trained to extract adverse events, leading to better performance. 

All our code and data are available as supplementary material.

%% file: figures/ade-example.tex
\begin{figure}[!tp]
\includegraphics[width=1.00\linewidth]{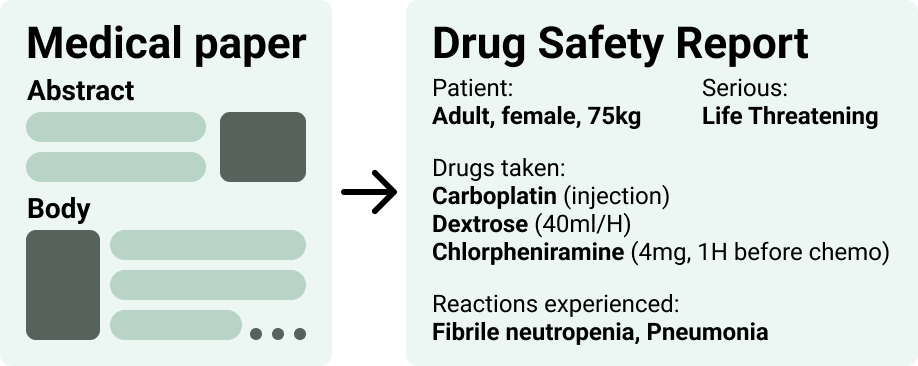} \centering
\caption{\OurDataset{} consists of 65k \PM{} abstracts and 19k full text papers, accompanied by 256k document-level drug safety reports. The schematic illustrates the core information that constitutes a drug safety report (they often contain much more detailed information as well). These reports are created by pharmacovigilance experts and are vital for drug safety monitoring.} \label{fig:ade-example}
\end{figure}

%% file: figures/article_report_trends.tex
\begin{figure}[t!]
\includegraphics[width=0.98\linewidth]{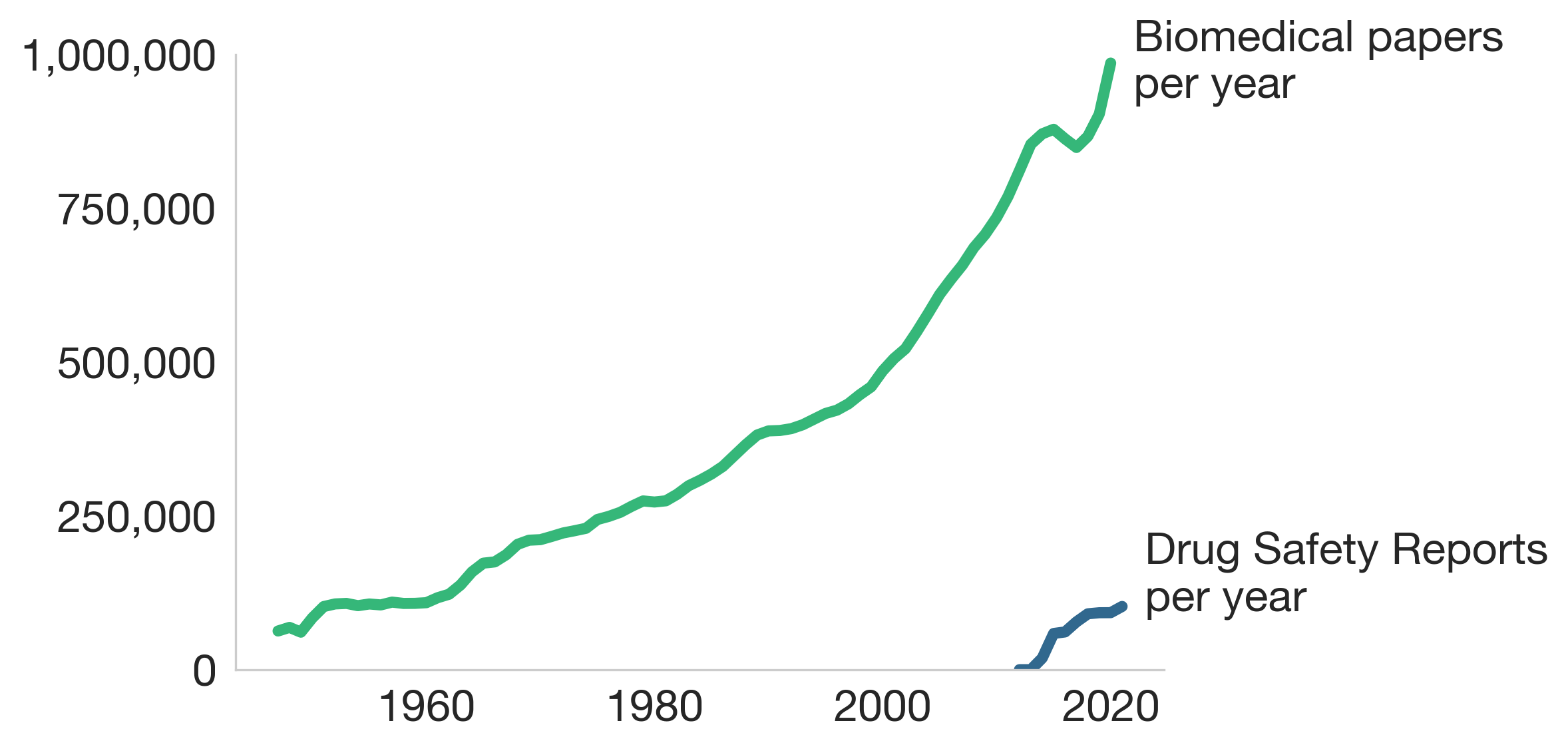}
\caption{The number of peer-reviewed biomedical papers published each year is accelerating (as indexed in Medline). The total number of drug safety reports originating from articles is on the rise as well, but the trend indicates stagnation (reports submitted to FAERS from 2012 onwards).} \label{fig:article-report-trend}
\end{figure}

%% file: content/reporting.tex
Pharmaceutical companies are required to participate in drug safety reporting for the drugs they produce. Regulations differ across regions of the world. In this work, we focus on the pharmacovigilance process as defined by U.S.\ regulations.

The reporting process starts with a PV literature review stage. Periodically, a vast database of biomedical literature is queried to retrieve new publications that could describe an adverse event with regard to a drug of interest. Conventionally this is done by matching the trade name of the drug or names of its active substances. These queries are designed by experts and depend on the specific use-case, but they always aim for wide coverage; there are strong regulatory fines associated with missing reports, which creates strong incentives for very high recall. Reports can also originate from other modalities such as forms, emails, and social media. In this work, we only focus on reports originating from biomedical publications.

Once a set of candidate publications is found, a triaging process begins. For example, papers that mention a serious adverse event should be prioritized, as these reports need to be submitted in a strict time window. This is often done via another high recall system that matches words such as `serious' and `life threatening' via a lexicon-based approach.

Each resulting publication is investigated by expert PV workers in a multi-stage pipeline, which can differ across companies. Typically, the initial flagging of potential ADEs is done by non-clinician PV workers. Evidence is flagged and can be mapped to a standardized ontology to introduce uniformity in downstream stages. Subsequently, clinicians review the report and refine the event details before the report is submitted.

In this work, we abstract away the details of this human-based workflow and model the task as taking in a biomedical publication and outputting the final pharmacovigilance report. Systems that perform well at this task could go a long way towards automating pharmacovigilance.

%% file: content/related-work.tex
\paragraph{Biomedical NLP}

LMs have pushed the frontiers of biomedical NLP. These models generally follow the Transformer architecture \citep{vaswani2017attention, devlin2018bert, radford2019language, brown2020language, raffel2020exploring, nori2023capabilities}. LMs, sometimes specifically tailored towards the biomedical domain, achieve state-of-the-art results across a range of biomedical benchmarks \citep{yasunaga2022linkbert,luo2022biogpt,singhal2022large}.
For example, LMs have achieved single-human performance on PubMedQA \citep{jin2019pubmedqa}, an expert-labeled biomedical question answering task with yes/no/maybe labels. Potentially, such models could be useful for PV as well. A key challenge is that PV requires processing entire biomedical publications, which PubMedQA does not support but \OurDataset{} does.

Recently, \citet{zhao2022pmc} introduced PMC-Patients, a large-scale dataset for patient-to-patient or patient-to-article retrieval built on top of \PM{}. 
\OurDataset{} can be seen as complementing this effort; instead of retrieving relevant papers, \OurDataset{} aims to extract structured patient information from biomedical publications for pharmacovigilance purposes. Both the extraction of the information as well as the retrieval of relevant articles are highly relevant for Evidence-Based Medicine (EBM; \citealt{sackett1997evidence}) and pharmacovigilance.

\paragraph{Adverse Drug Event Extraction} 
Previous work has focused on ADE extraction. However, almost all ADE datasets utilize some form of span-level annotations created by medical experts \citep{wallace2016extracting, roberts2017overview, nye2018corpus, kang2019pretraining, dirkson2022others}. This severely limits the scale of these approaches \citep{basile2019artificial}. \citet{nye2018corpus} annotate an impressive 5000 abstracts but in part utilize non-expert annotations. \citet{wallace2016extracting} combine a document-level resource for Randomized Control Trial reports with their supporting literature and use distant supervision to derive pseudo span-level labels. 

\OurDataset{} relies on the historical output of safety reporting in the U.S. Thus, it is orders of magnitude larger than these resources without requiring any additional expert labels, and it can automatically be expanded over time when new reports become available. This grounding in historical data entails that \OurDataset{} closely matches the real-world clinical and regulatory task of PV. In addition, since we consider adverse drug event extraction at the document-level, we circumvent the need for span-level labels.

\paragraph{FDA Adverse Event Reporting System} The FDA Adverse Event Reporting System (FAERS; \citealt{fda_faers}) is used as a cornerstone resource for drug safety research. Previous work has focused on pre-processing FAERS, which can include grounding drug and reaction mentions to medical ontologies and detecting duplicate reports \citep{banda2016curated, hauben2021more, khaleel2022standardized, 10.3389/fdsfr.2022.918897, hung2022more}. In contrast, \OurDataset\ is focused on improving the process of entering drug safety reports into FAERS, starting from the biomedical literature. 

\citet{xu2014large} combine both FAERS and biomedical literature for enhanced drug safety signal mining. We go one step further and explicitly link reports from FAERS with their originating documents, which allows us to create a document-level drug event extraction task.

%% file: content/data.tex
\subsection{Dataset Description}
Each entry of \OurDataset{} consists of one article and a list of associated reports. Articles and reports both contain many different features and metadata. In this section we limit ourselves to discussing only the most prominent features of our dataset. A full enumeration of all fields is given in \appref{app:schema} (for reports) and \appref{app:article-schema} (for articles).

\subsubsection{\PM{} Articles}
Each article contains a title and an abstract. If the full-text paper is openly accessible, it is also included together with its corresponding license. Articles also feature lists of keywords, Medical Subject Headings (MeSH; \citealt{lipscomb2000medical}), and a list of chemical substances mentioned in the publication.

The abstract and article metadata was parsed from the Medline distribution \citep{medline} using the \texttt{pubmed-parser} package \citep{Achakulvisut2020}. If available, the full-text paper was pulled from \PM{} Central Open Access Subset \citep{nlm_pmcos_2003}, using their provided API.\footnote{\url{https://www.ncbi.nlm.nih.gov/pmc/tools/openftlist/}}

\subsubsection{Drug Safety Reports}
A report contains clinically-relevant information about the described patient in the form of reported patient biological sex, weight, age group, and the age at which the event first occurred. Not all information is always present in the reports; this depends on what exactly the authors described in their article.

Each report features a list of drugs, each with their own set of fields. Every drug consists of one active ingredient. If available, the drug may feature additional details such as the product name of the drug, the drug administration route, the (cumulative) dosage taken, the action taken with this drug (e.g., dose increased), and whether the drug was considered a potential cause of the adverse reaction by the authors or not. If provided in the article, the reports can even describe the exact lot number of the drug product taken by the patient.

Each report also features a list of reactions. Each reaction is characterized by an entry from the standardized MedDRA  ontology (Medical Dictionary for Regulatory Activities; \citealt{brown1999medical}), as well as a field describing the outcome (e.g., recovered, recovering, fatal).

\subsection{Dataset Analysis}

\OurDataset{} features articles published between 1968 and 2022, with a stark increase in articles from 2013 onwards, corresponding to new PV-related legislation in Europe in 2012 \citep{fornasier2018historical}. \Figref{fig:article-dates} displays the article distribution starting from 2000. The associated reports all originate from a period between 2012 and 2022.
\input{figures/article_dates}

\OurDataset{} covers a broad range of topics. In total 55,951 unique article keywords are included. \figref{fig:top-keywords} shows the most prominent ones.
\input{figures/keyword-distribution.tex}

The median full-text paper in \OurDataset{} is about 20k characters long. \Tableref{table:length-distribution} displays the quartiles for both the abstract and full-text length in number of characters and tokens. We note that the average full-text paper is much longer than the context window used in many present-day LMs. 
\input{tables/length-distribution}

While \OurDataset{} is rooted in a U.S.-based resource, other countries are represented as well. \Figref{fig:occur-countries} illustrates from which countries the reports originated. Some regions are underrepresented, indicating an avenue for future work.
\input{figures/occur-country}

Not all report attributes are strictly required and thus show up across \OurDataset{} in varying frequencies. For example, the \texttt{patient~sex} attribute is present in 74.5\% of reports, while \texttt{patient~age~group} is only present in 17.9\% of reports. \appref{app:report-freq} outlines all attribute frequencies.

\subsection{Dataset Creation}
\OurDataset{} is created by matching articles parsed from Medline with drug safety reports entered in FAERS. To avoid ambiguity, we only consider articles with a unique \PM{} identifier and a unique title. Only reports containing an explicit reference to a supporting paper are considered. 

Unfortunately, this reference to the supporting literature is not structured. We parse the article title out of this unstructured reference. If we find a title that exactly matches a title in our set of articles, we enter both the article and associated report in \OurDataset. Otherwise, we drop the report.

When creating \OurDataset{}, we prioritized creating high-precision matches. Future work could expand the size of our dataset by considering a more sophisticated procedure to match articles and reports -- e.g., by using metadata other than the article titles.

%% file: figures/article_dates.tex
\begin{figure}[t]
\includegraphics[width=\linewidth]{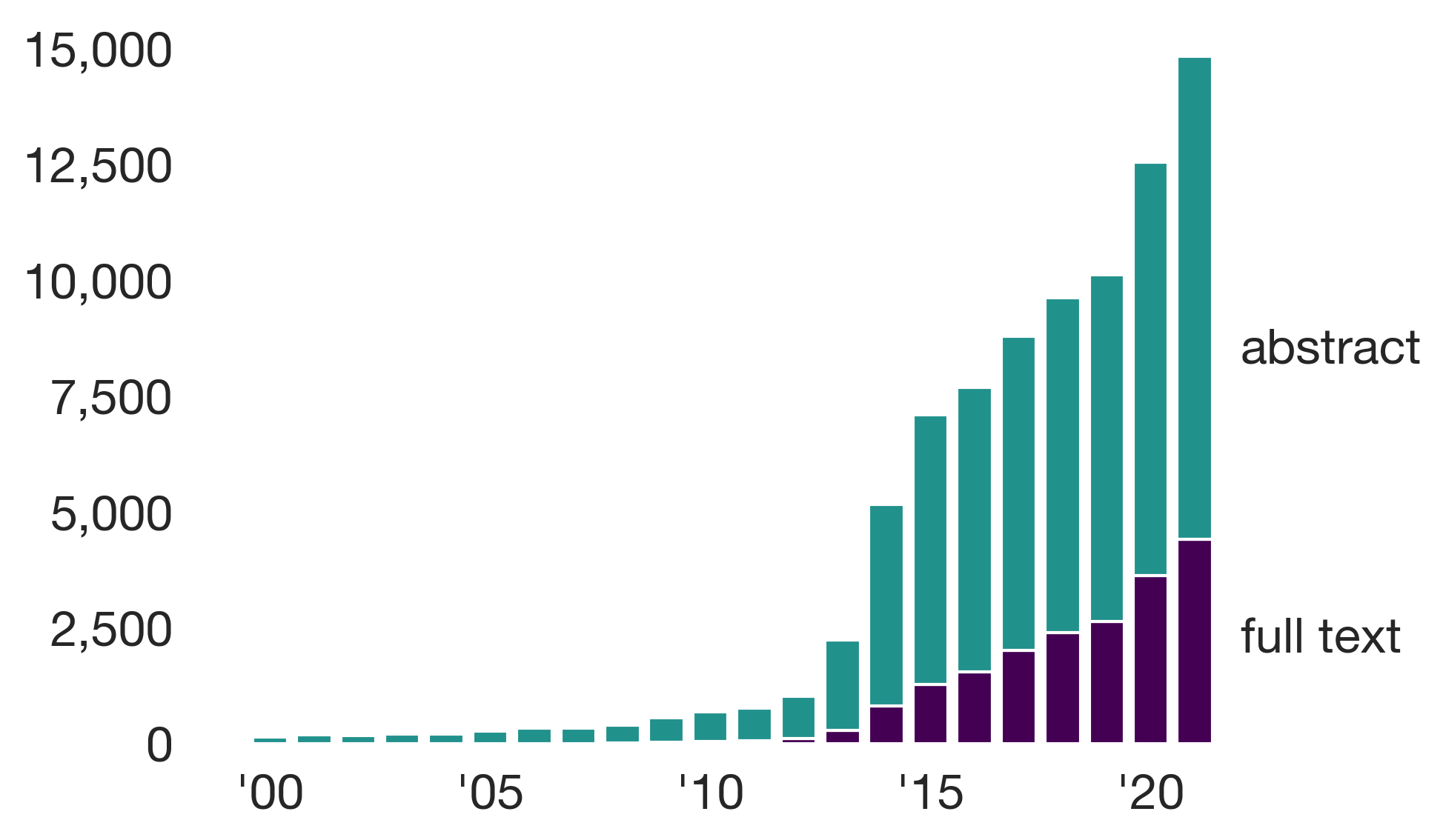}
\centering
\caption{
Number of {\OurDataset} abstracts and full-text papers published over time.
Articles published before 2000 are not visualized, there are only 1,519 of them.} 
\label{fig:article-dates}
\end{figure}

%% file: figures/keyword-distribution.tex
\begin{figure}[t]
\includegraphics[width=0.8\linewidth]{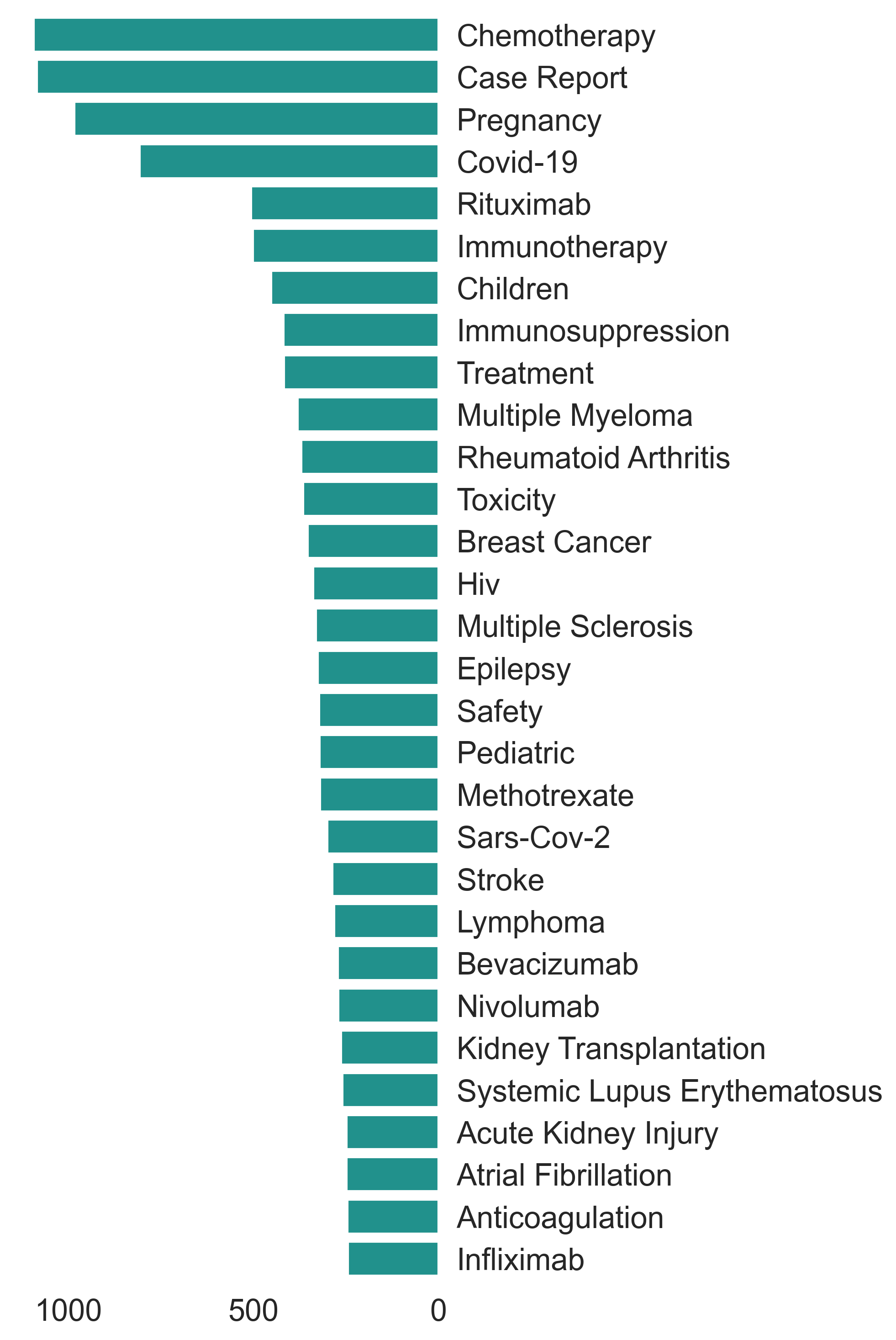} \centering
\caption{Number of occurrences for the 30 most frequent keywords in \OurDataset{} publications.}
\label{fig:top-keywords}
\end{figure}

%% file: tables/length-distribution.tex
\begin{table}[tp]
\begin{tabular}{@{}lrrr@{}}
\toprule
percentile:       & 25\textsuperscript{th}   & 50\textsuperscript{th}   & 75\textsuperscript{th}   \\ \midrule
abstract length  &     &   &   \\
\quad \# characters  & 825    & 1,263  & 1,679  \\
\quad \# tokens  & 177    & 275  & 383  \\
full-text length &  &  &  \\ 
\quad \# characters & 14,801 & 19,935 & 29,531 \\ 
\quad \# tokens & 3,761 & 5,152 & 7,890 \\  \bottomrule
\end{tabular} \centering
\caption{Abstract and full-text length percentiles of \OurDataset{} in number of characters and tokens. Tokenization done with the OpenAI's \texttt{tiktoken} package, using the vocabulary of the \texttt{text-davinci-002} model.}
\label{table:length-distribution}
\end{table}

%% file: figures/occur-country.tex
\begin{figure*}[t!]
\includegraphics[width=\textwidth]{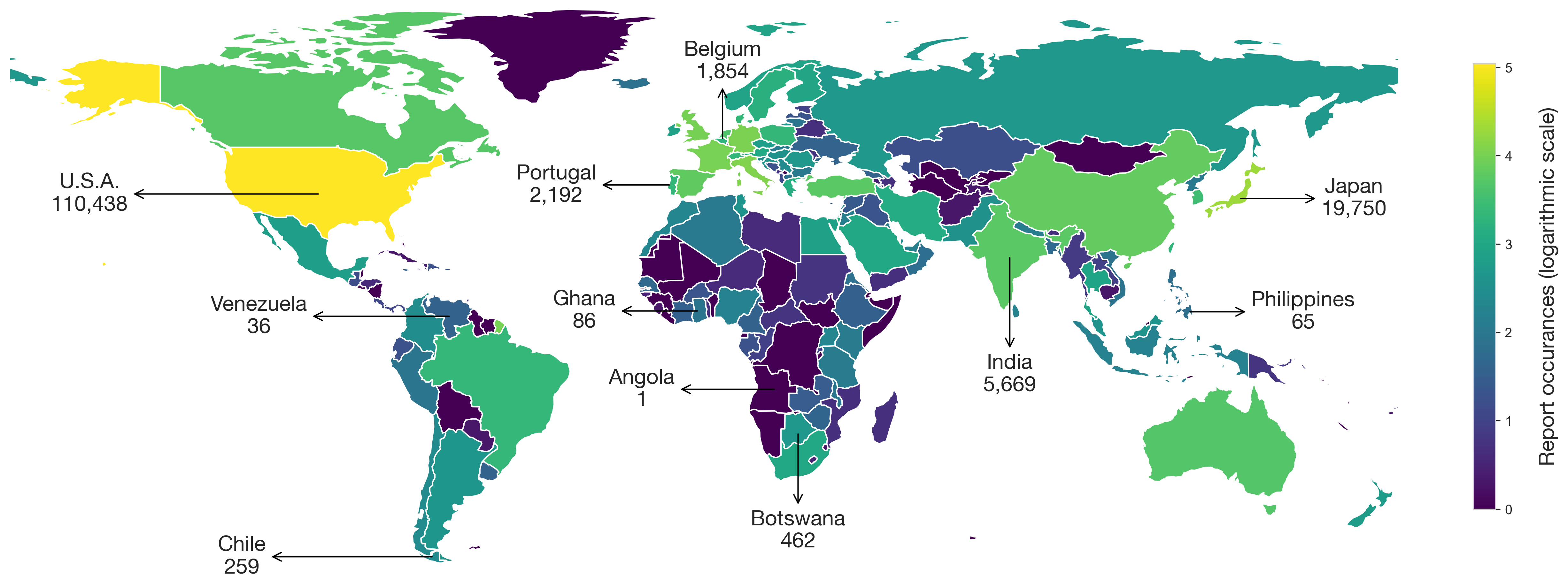}
\caption{Number of drug safety reports in \OurDataset{} originating from a given country. Colors follow a log scale. A selection of countries are specifically highlighted with their exact number of drug safety reports annotated.}
\label{fig:occur-countries}
\end{figure*}

%% file: content/metrics.tex
In this work, we focus on the task of predicting the core information of a report given a full-text paper, which we call Report-Extraction. Accurate and autonomous extraction of drug safety reports can have a large impact on PV by increasing the quality of safety signals and decreasing the time required to surface new signals.

\subsection{Core Reports}
We reduce the complexity of the detailed reports by only predicting the 4 core attributes:
\begin{enumerate}\setlength{\itemsep}{0pt}
    \item \texttt{Serious}: The seriousness of the adverse event. Equal to \texttt{1} if the adverse event resulted in death, a life threatening condition, hospitalization, disability, congenital anomaly, or any other serious condition. If none of the above occurred, equal to  \texttt{2}.
    \item \texttt{Patientsex}: The reported biological sex of the patient. \texttt{0} for unknown, \texttt{1} for male, \texttt{2} for female.
    \item \texttt{Drugs}: The set of all active substance names of the drugs discussed in the report. For example: \texttt{azathioprine, infliximab, mesalamine, prednisolone}.
    \item \texttt{Reactions}: The set of all reaction terms discussed in the report. For example: \texttt{Epstein-Barr virus infection reactivation, Idiopathic interstitial pneumonia}.
\end{enumerate}

For the Report-Extraction task, we only consider reports where all these 4 attributes are present. While \OurDataset{} reports contain more detailed attributes as well, we leave predicting these details as future work.

\subsection{The Report-Extraction Dataset}
We create a new dataset specifically for this task by manipulating \OurDataset{}. First, we restrict ourselves to only articles with a full-text paper available. Additionally, we only consider articles with less than 10 associated reports, since we found that the few articles with more were often very large survey papers discussing a broad range of adverse effects. If multiple reports per article are available, one report is sampled to act as the gold label of our task. We leave the task of predicting a variable number of reports per publication, which \OurDataset{} supports, as future work.

We divide the data into \texttt{train}/\texttt{test} splits by taking articles published before 2021 as training instances and the rest as testing instances. This adds a temporal generalization component to our task. Finally, we create a \texttt{validation} split by uniformly holding-out 20\% of the training samples. 

We deliberately created a test scenario that simulates the real-world situation these models will face: they will have been developed on data up to a specific time point and then, by necessity, they will encounter reports from later time periods. It is vital that we study how models behave in this challenging scenario.

The resulting dataset sizes and article dates are given in \tableref{table:icsr-summary-statistics}. We distribute this subset of our dataset in structured format as well. 

\input{tables/icsr-dataset-statistics}

\subsection{Report-Extraction Performance}
To estimate performance, we need to define a similarity metric between two core reports. This is achieved by taking a weighted average over the 4 attribute similarities.\footnote{The weight factors are $1/6$ for the \texttt{serious} and \texttt{patientsex} scores, and $1/3$ for the \texttt{drugs} and \texttt{reactions} scores.}
For \texttt{serious} and \texttt{patientsex}, the similarity is the conventional classification accuracy. For \texttt{drugs} and \texttt{reactions}, the set precision and recall metrics are used. Every predicted drug or reaction in these sets is either correct or wrong, based on an exact string match. We report the average of all the report-level F1 scores, calculated using the weighted attribute precision and recall scores. This is a strict metric, since multiple correct ways of describing the same drug or reaction are not taken into account. In future work, medical ontologies can be used to normalize drug and reaction mentions to create more lenient metrics.

\subsection{Inter-Annotator Agreement}
A single article can be linked to multiple reports. Often, these reports comment on the same underlying adverse event but were submitted by independent people or institutions. These situations can be used to estimate a lower-bound on the Inter-Annotator Agreement (IAA).

For every article with multiple reports available, we randomly validate one core report against another. Using our Report-Extraction Performance, this produces an IAA score of 72.04\% F1.

As a random baseline, we consider validating a core report against another report uniformly sampled from the entire dataset. This produces an F1 of 24.28\% and serves as lower bar for non-trivial performance. This score is significantly larger than 0\% mainly due to high random guessing accuracy on the \texttt{serious} and \texttt{patientsex} attributes.

%% file: tables/icsr-dataset-statistics.tex
\begin{table}[tp]
\centering
\setlength{\tabcolsep}{8pt}
\begin{tabular}{ llll }
\toprule
\multirow{2}{*}{split} & \multirow{2}{*}{size} & \multicolumn{2}{c}{article date} \\
                       &                             & min.                & max.               \\ \midrule
\texttt{train}                  & 9,624 (62\%)                      & 1990                & 2020               \\
\texttt{validation}            & 2,407 (15\%)                       & 1985                & 2020               \\
\texttt{test}                   & 3,628 (23\%)                      & 2021                & 2022               \\ \bottomrule
\end{tabular}
\centering
\caption{Sizes of the Report-Extraction splits and corresponding ranges of article publish dates.} \label{table:icsr-summary-statistics}
\end{table}

%% file: content/experiments.tex
Motivated by the recent success of LLMs, we choose to model Report-Extraction as a sequence-to-sequence problem.\footnote{Different views of the Report-Extraction task are possible. For example, it could be defined as a set of (multi-label) classification tasks.} Given a full-text paper as input, we train models to predict the core report in a stringified format, such as \texttt{``serious: 1  patientsex: 1 drugs: azathioprine, infliximab, mesalamine, prednisolone reactions: epstein-barr virus infection reactivation, idiopathic interstitial pneumonia''}.

We report \texttt{validation} results for all models considered. Only the best models are subsequently evaluated on the \texttt{test} split.

\subsection{Few-shot In-context Learning}
\input{tables/gpt-experiments}

First, we evaluate the few-shot in-context learning performance on our dataset achieved by OpenAI's \texttt{text-davinci-002}, \texttt{text-davinci-003}, \texttt{gpt-3.5-turbo}, and \texttt{gpt-4} models \citep{brown2020language}. A key limitation of in-context learning is that both the few-shot demonstrations and the actual input need to fit in the same context window. Given the average length of our inputs, the context window becomes a constraint: most of the full-text papers do not fit the \texttt{text-davinci-003} context window of 4,096 tokens (see \tableref{table:length-distribution}). 

Thus, we aim to maximally utilize the available context window. Given a fixed natural description prompt of the task (see \appref{app:prompt} for the full prompt), we investigate the trade-off between the number of tokens dedicated to in-context demonstrations and the number of tokens of the input paper. Since it is prohibitive to include entire papers, we use only the abstracts for the demonstrations and truncate the full-text input paper to maximally fill the context window. We use the \texttt{DSP} package to implement all experiments \citep{khattab2022demonstrate}.

\Tableref{table:gpt-experiments} summarizes the experiments. We find the optimal trade-off to consist of 7 abstract-level demonstrations, which results in incorporating around 1,660 tokens of the final paper. 

On the validation set, this achieves a performance of 45.78\% F1 for \texttt{text-davinci-002}, 50.44\% F1 for \texttt{text-davinci-003}, and 51.71\% F1 for \texttt{gpt-4}.\footnote{All default hyperparameter settings were used for the OpenAI API calls. To save on costs, we validate and test on the first 100 examples of the respective splits.} While this performance is certainly non-trivial, especially given only 7 labeled examples, it is far from expert-level. We explored using the context window of \texttt{gpt-4} beyond 4096 tokens, but found no improvements when further scaling the amount of demonstrations or the amount of paper input tokens. The cheaper \texttt{gpt-3.5-turbo} model performs sub-par and struggles to properly format its generations.

The best \texttt{text-davinci-003} and \texttt{gpt-4-0312} models achieve 50.60\% F1 and 53.11\% F1 on \texttt{test} respectively. We conclude that, at least in our standard use of the methods, few-shot learning achieves non-trivial but unsatisfactory performance on our Report-Extraction task. See \Appref{app:example-outputs} for 10 examples.

\subsection{Fine-tuned Models}

We further experiment with fine-tuning our own specialized models for the Report-Extraction task. We consider the suite of \texttt{FLAN-T5} models \citep{chung2022scaling}, which are based on the encoder-decoder Transformer architecture \citep{vaswani2017attention}. \Tableref{table:finetune-experiments} summarizes the experiments. 

The most successful run consisted of fine-tuning \texttt{FLAN-T5-Large} on a source context window of 2048 tokens and a target context window of 256 tokens. This achieves 62.28\% F1 on \texttt{validation}.

Given a fixed context window of 512 or 1,024 tokens, the larger \texttt{FLAN-T5-XL} model performs better. For a given model size, longer context windows improve performance. We leave the further scaling of model sizes and context windows as future work.

Models were trained for up to 5 epochs with a starting learning rate of $0.0001$, linearly scheduled. We used the Adafactor optimizer with default hyperparameters \citep{shazeer2018adafactor}.

We used greedy decoding to form the generations. Beam search decoding, with a beam width of 8, did not improve performance. Evaluating \texttt{FLAN-T5-Large} with 2048 source and 256 target tokens on the \texttt{test} split results in 59.1\% F1.

\input{tables/finetune-experiments}

\subsection{Attribute-level Performance}
Fine-tuned models have better Report-Extraction Performance compared to in-context learning models. In table \tableref{table:attribute-level}, we break down the performance of the best fine-tuned and in-context model on the \texttt{validation} split per predicted attribute. \texttt{FLAN-T5-Large} and \texttt{gpt-4} attain similar performance when predicting \texttt{seriousness} and \texttt{patientsex}. However, \texttt{FLAN-T5-Large} outperforms \texttt{gpt-4} at predicting the correct \texttt{drugs} and \texttt{reactions}. While these two attributes are hard to predict in general, we hypothesize the fine-tuned model attains better performance because it was able to learn from more reports during training, allowing it to better captured the specific terminology used in these reports as well as their prior distribution.

\input{tables/attribute-level}

%% file: tables/gpt-experiments.tex
\begin{table*}[tp]
\begin{tabular}{@{}l c c @{\hspace{10pt}}c@{\hspace{10pt}} c c c @{}}
\midrule
model                             & \# demos & \begin{tabular}[c]{@{}c@{}}\# input paper \\ tokens (avg)\end{tabular} & \begin{tabular}[c]{@{}c@{}} REP \\ (\% F1)\end{tabular} & \begin{tabular}[c]{@{}c@{}}Parse \\ percentage\end{tabular} & \begin{tabular}[c]{@{}c@{}}\# generation \\ tokens (avg)\end{tabular} & \begin{tabular}[c]{@{}l@{}}\# context \\ tokens (avg)\end{tabular} \\ \midrule
\texttt{text-davinci-002 }                 & 5        & 2347                        & 44.15       & 100                      & 41    & 3871                          \\
\texttt{text-davinci-002 }                 & 7        & 1669                        & 45.78       & 97                      & 35                           & 3956                          \\
\texttt{text-davinci-002 }                 & 10       & 845                         & 45.91       & 98                      & 43                         & 3965                          \\
\texttt{text-davinci-002 }                 & 12       & 385                         & 45.80       & 98                      & 36                         & 3968                          \\ \midrule
\texttt{text-davinci-003 }                  & 6        & 2070                        & 48.13       & 100                      & 50                         & 3968                          \\
\textbf{\texttt{text-davinci-003 }}                  & \textbf{7}        & \textbf{1669}                        & \textbf{50.45}     & \textbf{99}                      & \textbf{47}                         & \textbf{3968}                          \\
\texttt{text-davinci-003 }                  & 8        & 1440                        & 47.16       & 100                      & 54                         & 3959                          \\ \midrule
\texttt{gpt-3.5-turbo-0310}                & 7        & 1710                        & 30.55       & 76                    & 29                         & 3955                          \\ \midrule
\begin{tabular}[c]{@{}l@{}}\textbf{\texttt{gpt-4-0312}} \\ \quad(\textbf{4k context})\end{tabular} & \textbf{7}        & \textbf{1710}                        & \textbf{51.71}       & \textbf{100}                   & \textbf{43}                         & \textbf{3954}                          \\
\texttt{gpt-4-0312}                        & 7        & 3638                        & 49.69       & 100                   & 43                         & 5925                          \\
\texttt{gpt-4-0312}                        & 14       & 3151                        & 48.00       & 100                   & 38                         & 7215                          \\ \midrule
\end{tabular} \centering
\caption{Few-shot in-context learning results on the \OurDataset{} Report-Extraction task (\texttt{validation} split). For each model, we vary the combination of number of few-shot demos and the amount of tokens dedicated towards the input paper. REP denotes the Report-Extraction Performance. Parse percentage denotes the frequency of times the model formed a well-structured generation. Models were evaluated on the first 100 examples of \texttt{validation}.}
\label{table:gpt-experiments}
\end{table*}

%% file: tables/finetune-experiments.tex
\begin{table*}[tp]
\setlength{\tabcolsep}{8pt}
\begin{tabular}{@{} l c c @{\hspace{48pt}}c@{\hspace{48pt}}  c c @{}}
\toprule
model                                                            & \begin{tabular}[c]{@{}c@{}}\# source \\ tokens\end{tabular} & \begin{tabular}[c]{@{}c@{}}\# target \\ tokens\end{tabular} & \begin{tabular}[c]{@{}c@{}}REP\\ (\% F1)\end{tabular} & \begin{tabular}[c]{@{}c@{}}Parse \\ Percentage\end{tabular} & \begin{tabular}[c]{@{}c@{}}\# generation \\ tokens (avg)\end{tabular} \\ \midrule
\textbf{\texttt{FLAN-T5-Large}}                                                    & \textbf{2048}                                                        & \textbf{256}                                                                                                             & \textbf{62.28}                                                  & \textbf{98.96}                                                                 & \textbf{59.60}                                                                    \\
\texttt{FLAN-T5-Large}                                                    & 2048                                                        & 128                                                                                                               & 61.39                                                      & 99.58                                                                 & 52.96                                                                     \\
\texttt{FLAN-T5-Large}                                                    & 1024                                                        & 256                                                                                                              & 55.88                                                       & 96.05                                                                 & 75.08                                                                      \\
\texttt{FLAN-T5-Large}                                                    & \phantom{0}512                                                         & 128                                                                                                             &  50.92                                                     & 94.72                                                                 & 53.60                                                                      \\ \midrule
\texttt{FLAN-T5-XL}                                                       & 1024                                                        & 256                                                                                                              & 58.32                                                      & 99.46                                                                  & 48.82                                                                      \\
\texttt{FLAN-T5-XL}                                                       & \phantom{0}512                                                         & 256                                                                                                              & 53.19                                                      & 97.55                                                                 & 64.69                                                                      \\
\bottomrule
\end{tabular} \centering
\caption{Fine-tuning results on the \OurDataset{} Report-Extraction task (\texttt{validation} split). REP denotes the Report-Extraction Performance. Parse percentage denotes the frequency of well-structured model outputs.}
\label{table:finetune-experiments}
\end{table*}

%% file: tables/attribute-level.tex
\begin{table}[]
\centering
\begin{tabular}{@{}lll@{}}
\toprule
& \texttt{FLAN-T5-Large}     & \texttt{gpt-4-0312}    \\ 
& \% F1     & \% F1    \\ 
\midrule
\texttt{seriousness} & 92.90       & 94.00    \\
\texttt{patientsex}  & 92.65       & 93.00    \\
\texttt{drugs}       & 61.83       & 50.99    \\
\texttt{reactions}   & 34.13       & 12.62    \\ 
\bottomrule
\end{tabular}
\caption{Per attribute performance of the best \texttt{FLAN-T5-Large} and \texttt{gpt-4-0312} runs on the \texttt{validation} split.}
\label{table:attribute-level}
\end{table}

%% file: content/results.tex
\input{figures/classification}

Our primary goal is to improve the scalability and accuracy of PV using NLP. The above experiments highlighted the potential for LMs to autonomously fill in ADE reports. However, fully autonomous drug event reporting systems are unlikely to achieve widespread adoption \emph{today}. Mainly because of the challenging nature of this task and the high cost of errors, human experts will remain vital for effective solutions in the years to come.

However, our models can still deliver tangible value by augmenting existing expert-based workflows. Given the vast number of biomedical papers published, it is increasingly impractical to thoroughly vet every candidate publication (as is currently being done). Drug manufacturers are looking to more efficiently triage the literature to prioritize efforts, as this minimizes risk from regulatory fines and maximizes public safety.

Such a triaging system is typically based on a naive lookup: finding all papers that match a drug name is likely to find all papers where that drug engages in an adverse event. Unfortunately, such a system has low precision, causing human effort to be wasted investigating irrelevant papers.

We find that our model predictions achieve a higher performance at finding adverse events concerning specific reactions, compared to the lookup baseline. We measure this through the macro average F1 score on the binary classification task of predicting per paper if a reaction was part of an adverse events. \Figref{fig:classification} shows the results for the 30 most frequent reactions in the \texttt{validation} split. High-recall baselines still have a valuable place in the PV review process, but our system could be used to more efficiently prioritize effort. \Appref{app:drug-classification} describes the same experiment for drugs.

Future work could utilize all details of \OurDataset{} reports or incorporate the 65k abstract-level datapoints during training to  further improve utility for PV. For example, \OurDataset{} would support fine-tuning a question answering model for PV.

%% file: figures/classification.tex
\begin{figure}[t!]
\includegraphics[height=11cm]{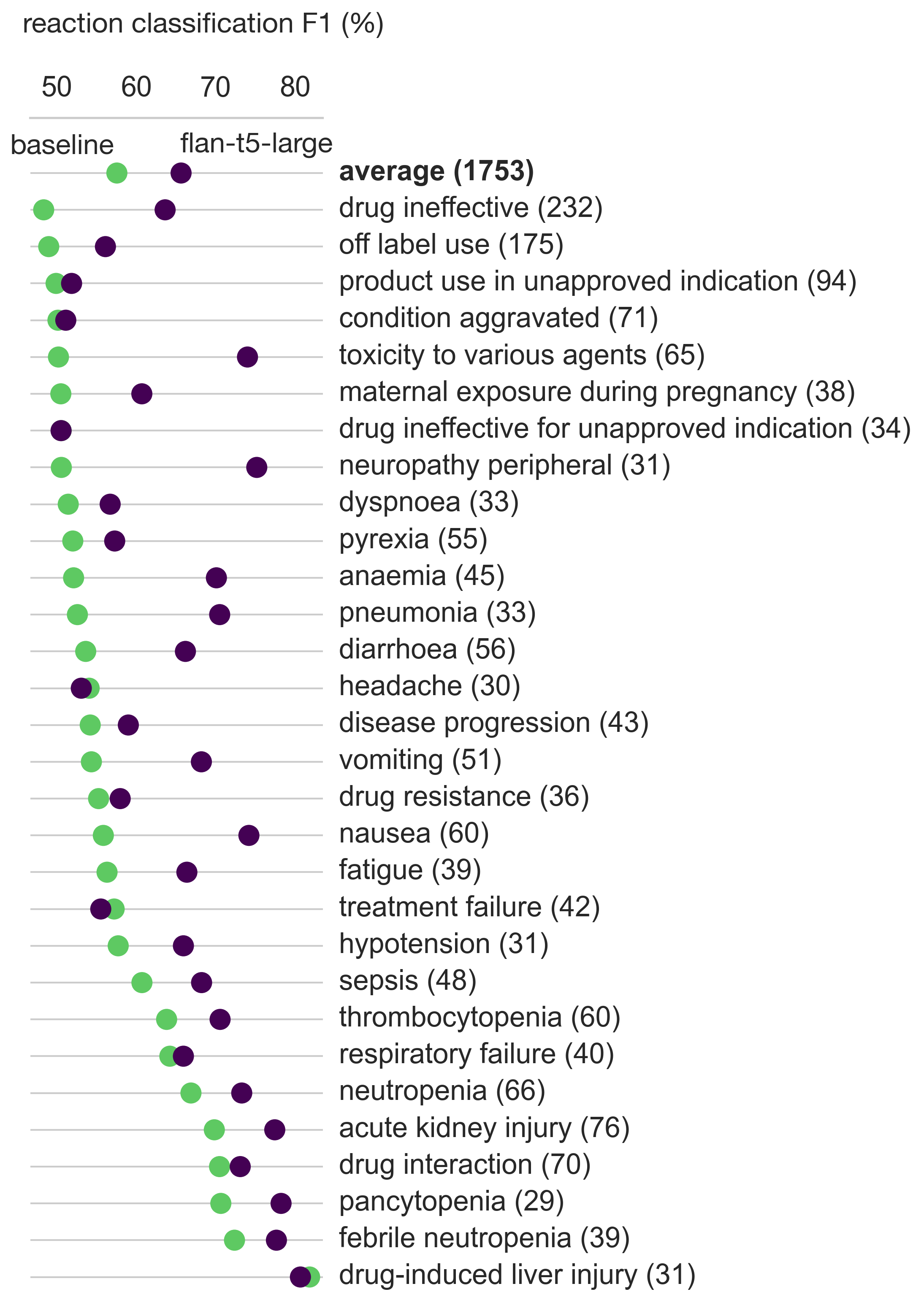}
\caption{Reaction classification performance across the 30 most frequent reactions in the \OurDataset{} \texttt{validation} set. Baseline performance in lighter color, \texttt{FLAN-T5} in darker color. Support in parentheses. Average performance in bold. Reactions are sorted by baseline performance.} \label{fig:classification}
\end{figure}

%% file: content/conclusion.tex
We introduced \OurDataset{}, a large-scale document-level Biomedical adverse Drug Event Extraction dataset. \OurDataset{} covers an important and challenging real-world task: extracting detailed drug safety reports from full-text biomedical publications. We find that LLMs struggle to get traction on this task using in-context learning. Fine-tuned models are more successful, but expert-level performance remains elusive. Nevertheless, our models have the potential to make drug safety research more efficient, and we demonstrated their utility in a conventional PV use-case. We release all data and models. We hope that \OurDataset{} stimulates new research in the high-impact area of drug safety monitoring.

%% file: content/limitations.tex
Drug Safety Reporting is an important real-world task. Submitting faulty reports or consistently underreporting specific adverse events could have profound impacts for public safety. LMs are known to make mistakes and fabricate evidence, they are almost invariably biased towards specific predictions, and they can be prone to adversarial attacks \citep{bender2021dangers, zhang2020adversarial}. Thus, the resources put forth in this paper should not be naively applied to automate safety reporting. Rather, we suggest that these systems could be integrated as an additional tool at the disposal of PV workers, and we encourage careful study of how to best empower these experts to work more efficiently and effectively. 

Different countries can face different health issues. When we develop biomedical language systems, it is important they work for everyone. Some countries are underrepresented in our dataset. Subsequent data collection efforts should focus on these countries to alleviate this issue. Additionally, confounders such as patient age and patient sex need to be taken into account to ensure satisfactory performance across different demographics.

%% file: content/acknowledgements.tex
We thank our anonymous reviewers for their insightful comments and suggestions. KD gratefully acknowledges funding from the FWO Fundamental Research PhD Fellowship (11632223N). KZ gratefully acknowledges funding from the Innovation Fund Denmark (IFD) through the Grand Solution project Hospital@Night.

%% file: appendix/schema.tex
The following paragraph enumerates the fields present in the drug safety reports and lists possible values if defined. It was adapted from the official description\footnote{\url{https://open.fda.gov/apis/drug/event/searchable-fields/}} of the FAERS fields found on OpenFDA \citep{kass2016openfda}. The dot in field names denotes nesting.
\\
\input{tables/schema}

%% file: tables/schema.tex
\noindent \textbf{companynumb}
Identifier for the company providing the report. This is self{-}assigned.

\noindent \textbf{fulfillexpeditecriteria}
Identifies expedited reports (those that were processed within 15 days).
Possible values: 1: True, 2: False

\noindent \textbf{occurcountry}
The name of the country where the event occurred.
Possible values: name: Country codes, link: http://data.okfn.org/data/core/country{-}list

\noindent \textbf{patient.drug.items.actiondrug}
Actions taken with the drug.
Possible values: 1: Drug withdrawn, 2: Dose reduced, 3: Dose increased, 4: Dose not changed, 5: Unknown, 6: Not applicable

\noindent \textbf{patient.drug.items.activesubstance.activesubstancename}
Product active ingredient, which may be different than other drug identifiers (when provided).

\noindent \textbf{patient.drug.items.drugadditional}
Dechallenge outcome information—whether the event abated after product use stopped or the dose was reduced. Only present when this was attempted and the data was provided.
Possible values: 1: Yes, 2: No, 3: Does not apply

\noindent \textbf{patient.drug.items.drugadministrationroute}
The drug’s route of administration.
Possible values: 001: Auricular (otic), 002: Buccal, 003: Cutaneous, 004: Dental, 005: Endocervical, 006: Endosinusial, 007: Endotracheal, 008: Epidural, 009: Extra{-}amniotic, 010: Hemodialysis, 011: Intra corpus cavernosum, 012: Intra{-}amniotic, 013: Intra{-}arterial, 014: Intra{-}articular, 015: Intra{-}uterine, 016: Intracardiac, 017: Intracavernous, 018: Intracerebral, 019: Intracervical, 020: Intracisternal, 021: Intracorneal, 022: Intracoronary, 023: Intradermal, 024: Intradiscal (intraspinal), 025: Intrahepatic, 026: Intralesional, 027: Intralymphatic, 028: Intramedullar (bone marrow), 029: Intrameningeal, 030: Intramuscular, 031: Intraocular, 032: Intrapericardial, 033: Intraperitoneal, 034: Intrapleural, 035: Intrasynovial, 036: Intratumor, 037: Intrathecal, 038: Intrathoracic, 039: Intratracheal, 040: Intravenous bolus, 041: Intravenous drip, 042: Intravenous (not otherwise specified), 043: Intravesical, 044: Iontophoresis, 045: Nasal, 046: Occlusive dressing technique, 047: Ophthalmic, 048: Oral, 049: Oropharingeal, 050: Other, 051: Parenteral, 052: Periarticular, 053: Perineural, 054: Rectal, 055: Respiratory (inhalation), 056: Retrobulbar, 057: Sunconjunctival, 058: Subcutaneous, 059: Subdermal, 060: Sublingual, 061: Topical, 062: Transdermal, 063: Transmammary, 064: Transplacental, 065: Unknown, 066: Urethral, 067: Vaginal

\noindent \textbf{patient.drug.items.drugauthorizationnumb}
Drug authorization or application number (NDA or ANDA), if provided.

\noindent \textbf{patient.drug.items.drugbatchnumb}
Drug product lot number, if provided.

\noindent \textbf{patient.drug.items.drugcharacterization}
Reported role of the drug in the adverse event report. These values are not validated by FDA.
Possible values: 1: Suspect (the drug was considered by the reporter to be the cause), 2: Concomitant (the drug was reported as being taken along with the suspect drug), 3: Interacting (the drug was considered by the reporter to have interacted with the suspect drug)

\noindent \textbf{patient.drug.items.drugcumulativedosagenumb}
The cumulative dose taken until the first reaction was experienced, if provided.

\noindent \textbf{patient.drug.items.drugcumulativedosageunit}
The unit for `drugcumulativedosagenumb`.
Possible values: 001: kg (kilograms), 002: g (grams), 003: mg (milligrams), 004: µg (micrograms)

\noindent \textbf{patient.drug.items.drugdosageform}
The drug’s dosage form. There is no standard, but values may include terms like `tablet` or `solution for injection`.

\noindent \textbf{patient.drug.items.drugdosagetext}
Additional detail about the dosage taken. Frequently unknown, but occasionally including information like a brief textual description of the schedule of administration.

\noindent \textbf{patient.drug.items.drugenddate}
Date the patient stopped taking the drug.

\noindent \textbf{patient.drug.items.drugenddateformat}
Encoding format of the field `drugenddateformat`. Always set to `102` (YYYYMMDD).

\noindent \textbf{patient.drug.items.drugindication}
Indication for the drug’s use.

\noindent \textbf{patient.drug.items.drugintervaldosagedefinition}
The unit for the interval in the field `drugintervaldosageunitnumb.`
Possible values: 801: Year, 802: Month, 803: Week, 804: Day, 805: Hour, 806: Minute, 807: Trimester, 810: Cyclical, 811: Trimester, 812: As necessary, 813: Total

\noindent \textbf{patient.drug.items.drugintervaldosageunitnumb}
Number of units in the field `drugintervaldosagedefinition`.

\noindent \textbf{patient.drug.items.drugrecurreadministration}
Whether the reaction occured after readministration of the drug.
Possible values: 1: Yes, 2: No, 3: Unknown

\noindent \textbf{patient.drug.items.drugrecurrence\linebreak.drugrecuractionmeddraversion}
The version of MedDRA from which the term in `drugrecuraction` is drawn.

\noindent \textbf{patient.drug.items.drugseparatedosagenumb}
The number of separate doses that were administered.

\noindent \textbf{patient.drug.items.drugstartdate}
Date the patient began taking the drug.

\noindent \textbf{patient.drug.items.drugstartdateformat}
Encoding format of the field `drugstartdate`. Always set to `102` (YYYYMMDD).

\noindent \textbf{patient.drug.items.drugstructuredosagenumb}
The number portion of a dosage; when combined with `drugstructuredosageunit` the complete dosage information is represented. For example, *300* in `300 mg`.

\noindent \textbf{patient.drug.items.drugstructuredosageunit}
The unit for the field `drugstructuredosagenumb'. For example, *mg* in `300 mg'.
Possible values: 001: kg (kilograms), 002: g (grams), 003: mg (milligrams), 004: µg (micrograms)

\noindent \textbf{patient.drug.items.drugtreatmentduration}
The interval of the field `drugtreatmentdurationunit' for which the patient was taking the drug.

\noindent \textbf{patient.drug.items.drugtreatmentdurationunit}
None
Possible values: 801: Year, 802: Month, 803: Week, 804: Day, 805: Hour, 806: Minute

\noindent \textbf{patient.drug.items.medicinalproduct}
Drug name. This may be the valid trade name of the product (such as `ADVIL` or `ALEVE`) or the generic name (such as `IBUPROFEN'). This field is not systematically normalized. It may contain misspellings or idiosyncratic descriptions of drugs, such as combination products such as those used for birth control.

\noindent \textbf{patient.patientagegroup}
Populated with Patient Age Group code.
Possible values: 1: Neonate, 2: Infant, 3: Child, 4: Adolescent, 5: Adult, 6: Elderly

\noindent \textbf{patient.patientonsetage}
Age of the patient when the event first occured.

\noindent \textbf{patient.patientonsetageunit}
The unit for the interval in the field `patientonsetage.`
Possible values: 800: Decade, 801: Year, 802: Month, 803: Week, 804: Day, 805: Hour

\noindent \textbf{patient.patientsex}
The sex of the patient.
Possible values: 0: Unknown, 1: Male, 2: Female

\noindent \textbf{patient.patientweight}
The patient weight, in kg (kilograms).

\noindent \textbf{patient.reaction.items.reactionmeddrapt}
Patient reaction, as a MedDRA term. Note that these terms are encoded in British English. For instance, diarrhea is spelled `diarrohea`. MedDRA is a standardized medical terminology.
Possible values: name: MedDRA, link: http://www.fda.gov/ForIndustry/DataStandards/\linebreak StructuredProductLabeling/ucm162038.htm

\noindent \textbf{patient.reaction.items.reactionmeddraversionpt}
The version of MedDRA from which the term in `reactionmeddrapt` is drawn.

\noindent \textbf{patient.reaction.items.reactionoutcome}
Outcome of the reaction in `reactionmeddrapt` at the time of last observation.
Possible values: 1: Recovered/resolved, 2: Recovering/resolving, 3: Not recovered/not resolved, 4: Recovered/resolved with sequelae (consequent health issues), 5: Fatal, 6: Unknown

\noindent \textbf{patient.summary.narrativeincludeclinical}
Populated with Case Event Date, when available; does `NOT` include Case Narrative.

\noindent \textbf{primarysource.literaturereference}
Populated with the Literature Reference information, when available.

\noindent \textbf{primarysource.qualification}
Category of individual who submitted the report.
Possible values: 1: Physician, 2: Pharmacist, 3: Other health professional, 4: Lawyer, 5: Consumer or non{-}health professional

\noindent \textbf{primarysource.reportercountry}
Country from which the report was submitted.

\noindent \textbf{primarysourcecountry}
Country of the reporter of the event.
Possible values: name: Country codes, link: http://data.okfn.org/data/core/country{-}list

\noindent \textbf{receiptdate}
Date that the \_most recent\_ information in the report was received by FDA.

\noindent \textbf{receivedate}
Date that the report was \_first\_ received by FDA. If this report has multiple versions, this will be the date the first version was received by FDA.

\noindent \textbf{receiver.receiverorganization}
Name of the organization receiving the report. Because FDA received the report, the value is always `FDA'.

\noindent \textbf{receiver.receivertype}
The type of organization receiving the report. The value,`6', is only specified if it is `other', otherwise it is left blank.
Possible values: 6: Other

\noindent \textbf{reporttype}
Code indicating the circumstances under which the report was generated.
Possible values: 1: Spontaneous, 2: Report from study, 3: Other, 4: Not available to sender (unknown)

\noindent \textbf{safetyreportid}
The 8{-}digit Safety Report ID number, also known as the case report number or case ID. The first 7 digits (before the hyphen) identify an individual report and the last digit (after the hyphen) is a checksum. This field can be used to identify or find a specific adverse event report.

\noindent \textbf{safetyreportversion}
The version number of the `safetyreportid'. Multiple versions of the same report may exist, it is generally best to only count the latest report and disregard others. openFDA will only return the latest version of a report.

\noindent \textbf{sender.senderorganization}
Name of the organization sending the report. Because FDA is providing these reports to you, the value is always `FDA{-}Public Use.`

\noindent \textbf{sender.sendertype}
The name of the organization sending the report. Because FDA is providing these reports to you, the value is always `2`.
Possible values: 2: Regulatory authority

\noindent \textbf{serious}
Seriousness of the adverse event.
Possible values: 1: The adverse event resulted in death, a life threatening condition, hospitalization, disability, congenital anomaly, or other serious condition, 2: The adverse event did not result in any of the above

\noindent \textbf{seriousnesscongenitalanomali}
This value is `1` if the adverse event resulted in a congenital anomaly, and absent otherwise.

\noindent \textbf{seriousnessdeath}
This value is `1` if the adverse event resulted in death, and absent otherwise.

\noindent \textbf{seriousnessdisabling}
This value is `1` if the adverse event resulted in disability, and absent otherwise.

\noindent \textbf{seriousnesshospitalization}
This value is `1` if the adverse event resulted in a hospitalization, and absent otherwise.

\noindent \textbf{seriousnesslifethreatening}
This value is `1` if the adverse event resulted in a life threatening condition, and absent otherwise.

\noindent \textbf{seriousnessother}
This value is `1` if the adverse event resulted in some other serious condition, and absent otherwise.

\noindent \textbf{transmissiondate}
Date that the record was created. This may be earlier than the date the record was received by the FDA.

%% file: appendix/article-schema.tex
The following paragraph enumerates the fields present in the articles. It was adapted from the \texttt{pubmed-parser} \citep{Achakulvisut2020} documentation.
\\

\input{tables/article-schema}

%% file: tables/article-schema.tex
\noindent \textbf{title}
Title of the article.

\noindent \textbf{pmid}
PubMed ID.

\noindent \textbf{issue}
The Issue of the journal.

\noindent \textbf{pages}
Pages of the article in the journal publication.

\noindent \textbf{abstract}
Abstract of the article.

\noindent \textbf{fulltext}
The full text associated with the article from the PubMed Central Open Access Subset, if available.

\noindent \textbf{fulltext\_license}
The license associated with the full text paper from the PubMed Central Open Access Subset, if available.

\noindent \textbf{journal}
Journal of the given paper.

\noindent \textbf{authors}
Authors, each separated by `;'.

\noindent \textbf{affiliations}
The affiliations of the authors.

\noindent \textbf{pubdate}
Publication date. Defaults to year information only.

\noindent \textbf{doi}
DOI.

\noindent \textbf{medline\_ta}
Abbreviation of the journal name.

\noindent \textbf{nlm\_unique\_id}
NLM unique identification.

\noindent \textbf{issn\_linking}
ISSN linkage, typically use to link with Web of Science dataset.

\noindent \textbf{country}
Country extracted from journal information field.

\noindent \textbf{mesh\_terms}
List of MeSH terms with corresponding MeSH ID, each separated by `;' e.g. `D000161:Acoustic Stimulation; D000328:Adult; ...' .

\noindent \textbf{publication\_types}
List of publication type list each separated by `;' e.g. `D016428:Journal Article'.

\noindent \textbf{chemical\_list}
List of chemical terms, each separated by `;'.

\noindent \textbf{keywords}
List of keywords, each separated by `;'.

\noindent \textbf{reference}
String of PMID each separated by `;' or list of references made to the article.

\noindent \textbf{delete}
Boolean, `False' means paper got updated so you might have two.

\noindent \textbf{pmc}
PubMed Central ID.

\noindent \textbf{other\_id}
Other IDs found, each separated by `;'.

%% file: appendix/report-freq.tex
\tableref{table:report-freq} contains the frequency of occurrence for the report attributes in \OurDataset{}.

\input{tables/report-freq}

%% file: tables/report-freq.tex
\begin{table}[]
\centering
\begin{tabular}{@{}ll@{}}
\toprule
attribute                    & frequency (\%) \\ \midrule
patientagegroup              & 17.94     \\
patientonsetage              & 68.94     \\
patientonsetageunit          & 68.94     \\
patientsex                   & 74.58     \\
patientweight                & 4.78      \\
summary                      & 10.07     \\
drugadministrationroute      & 74.32     \\
drugbatchnumb                & 10.94     \\
drugcumulativedosagenumb     & 0.40      \\
drugcumulativedosageunit     & 0.35      \\
drugenddate                  & 2.73      \\
drugenddateformat            & 2.73      \\
drugintervaldosagedefinition & 13.93     \\
drugintervaldosageunitnumb   & 13.93     \\
drugrecurreadministration    & 17.07     \\
drugseparatedosagenumb       & 13.77     \\
drugstartdate                & 6.15      \\
drugstartdateformat          & 6.15      \\
drugtreatmentduration        & 0.64      \\
drugtreatmentdurationunit    & 0.64      \\
drugrecurrence               & 0.63      \\
drugdosageform               & 17.78     \\
drugdosagetext               & 59.83     \\
drugstructuredosagenumb      & 29.33     \\
drugstructuredosageunit      & 29.33     \\
drugauthorizationnumb        & 31.98     \\
actiondrug                   & 76.79     \\
drugadditional               & 41.12     \\
drugindication               & 81.85     \\ 
activesubstance              & 99.32     \\
reactionoutcome              & 98.10     
\end{tabular}
\caption{Frequency of occurrence of each attribute per report. Attributes not mentioned have an occurrence of 100\%.}
\label{table:report-freq}
\end{table}

%% file: tables/prompt.tex
\noindent \textbf{Few-Shot Prompt:} \newline
Read a biomedical paper and extract information about the adverse drug event mentioned by the authors. Return a serious value ('1' for serious, '2' for not serious). Return a patientsex value ('1' for male, '2' for female). Return a list of drugs taken and reactions experienced.\newline%
\newline%
{-}{-}{-}\newline%
\newline%
Follow the following format.\newline%
\newline%
Question: \$\{What adverse drug event was described in the following context?\}\newline%
Context: \$\{biomedical paper that describes adverse drug events\}\newline%
Answer: \$\{the adverse drug event described in the context\}\newline%
\newline%
{-}{-}{-}\newline%
\newline%
Question: What adverse drug event was described in the following context?\newline%
Context: we report the case of a patient with b{-}cell prolymphocytic leukemia who was successfully treated with the novel humanized monoclonal antibody obinutuzumab. this patient was previously treated with the combination of rituximab and bendamustine and had recurrent infusion reactions. her treatment with rituximab and bendamustine was discontinued when she developed disease progression after 3 cycles of therapy. she was then treated with obinutuzumab 1000 mg on day 1 of every cycle and chlorambucil 0.5 mg/kg on days 1 and 15 every 28 days to which she had greater tolerability. after 4 cycles of treatment, she had resolution of her clinical symptoms, massive splenomegaly, and normalization of her white blood cell count.\newline%
Answer: serious: 1 patientsex: 2 drugs: bendamustine hydrochloride, rituximab reactions: cytopenia, treatment failure\newline%
\newline%
Question: What adverse drug event was described in the following context?\newline%
Context: sarcoid associated pulmonary hypertension (saph) is a common complication of sarcoidosis and is associated with poor prognosis. saph can be due to multiple synergistic mechanisms and current therapeutic strategies treat systemic sarcoidosis and pulmonary hypertension separately. several studies have been performed to develop an effective therapy for saph but have been met with mixed results. the ambition trial successfully treated incident patients with pulmonary arterial hypertension (pah) with the upfront combination of ambrisentan and tadalafil; however combination therapy has not yet been studied in patients with saph. here we report a cohort of patients with newly diagnosed saph who were treated with upfront combination therapy per the ambition study protocol. we report three subjects with newly diagnosed saph who were treated with combination ambrisentan and tadalafil. baseline hemodynamics were compared with those from surveillance right heart catheterization while on therapy. mean follow up period was 17 months. each subject demonstrated clinical and hemodynamic improvement with combination therapy. this series is the first to evaluate upfront combination ambrisentan and tadalafil therapy for treatment of newly diagnosed saph. despite the impressive clinical and hemodynamic improvement, the study is limited by its small size and retrospective nature. while these initial results are promising, further work is needed to fully evaluate this regimen for treatment of saph. (sarcoidosis vasc diffuse lung dis 2020; 37 (2): 234{-}238).\newline%
Answer: serious: 1 patientsex: 2 drugs: ambrisentan, infliximab, methotrexate, prednisolone, tadalafil reactions: off label use, urosepsis\newline%
\newline%
Question: What adverse drug event was described in the following context?\newline%
Context: haloperidol is a typical antipsychotic drug. this drug is still widely used in emergency medicine, psychiatry, and general medicine departments. it is mostly used for acute confusional state, psychotic disorders, agitation, delirium, and aggressive behaviour. overdose of haloperidol can cause sudden deaths. cardiopulmonary arrest related to use of haloperidol had been reported in literature as case reports but are very few. no such cases have been reported in india till now. we report a case of cardiac arrest due to the use of haloperidol.\newline%
Answer: serious: 1 patientsex: 1 drugs: haloperidol lactate reactions: cardiac arrest, ventricular tachycardia\newline%
\newline%
Question: What adverse drug event was described in the following context?\newline%
Context: neonatal nonoliguric hyperkalemia (nohk) is a metabolic abnormality that occurs in extremely premature neonates at approximately 24~h after birth and is mainly due to the immature functioning of the sodium (na+)/potassium (k+) pump. magnesium sulfate is frequently used in obstetrical practice to prevent preterm labor and to treat preeclampsia; this medication can also cause hypermagnesemia and hyperkalemia by a mechanism that is different from that of nohk. herein, we report the first case of very early{-}onset neonatal hyperkalemia induced by maternal hypermagnesemia.    a neonate born at 32~weeks of gestation developed hyperkalemia (k+ 6.4~mmol/l) 2~h after birth. the neonate's blood potassium concentration reached 7.0~mmol/l 4~h after birth, despite good urine output. the neonate and his mother had severe hypermagnesemia caused by intravenous infusion of magnesium sulfate given for tocolysis due to pre{-}term labor.    the early{-}onset hyperkalemia may have been caused by the accumulation of potassium ions transported through the placenta, the shift of potassium ions from the intracellular to the extracellular space in the infant due to the malfunctioning of the na+/k+ pump and the inhibition of renal distal tube potassium ion secretion, there is a possibility that these mechanisms were induced by maternal and fetal hypermagnesemia after maternal magnesium sulfate administration. because neonatal hyperkalemia poses a significant risk for the development of life{-}threatening cardiac arrhythmia, this case highlights the necessity of maternal blood magnesium monitoring during magnesium sulfate administration and neonatal blood potassium monitoring when there is severe maternal hypermagnesemia at delivery.\newline%
Answer: serious: 1 patientsex: 2 drugs: magnesium sulfate reactions: exposure during pregnancy, hypermagnesaemia, hypocalcaemia, hypotonia, product use in unapproved indication\newline%
\newline%
Question: What adverse drug event was described in the following context?\newline%
Context: doxycycline and minocycline are tetracyclines with the potential to cause hepatoxicity. although autoimmune{-}like hepatitis from minocycline is well{-}described, doxycycline{-}induced autoimmune hepatitis (diah) has only been described once. we report a rare case of diah with elevated liver enzymes over 5 times the normal upper limit, elevated immunoglobulin g, and high titers of antismooth muscle antibody and antinuclear antibody. by stopping doxycycline, our patient's liver enzymes normalized and immunoglobulin g and autoantibody titers rapidly downtrended. as long{-}term doxycycline therapy becomes more prevalent to treat acne vulgaris and other skin conditions, diah may become more prevalent and recognized.\newline%
Answer: serious: 1 patientsex: 2 drugs: doxycycline hyclate reactions: autoimmune hepatitis\newline%
\newline%
Question: What adverse drug event was described in the following context?\newline%
Context: oral mucositis, the most common adverse effect of radiotherapy (rt) and/or chemotherapy is observed in almost 97\% of patients with head and neck cancer. although several agents like corticosteroids, lidocaine and vitamins are available for its prevention or management, results are often disappointing. here we report on the effects of a topically applied, highly purified natural deoxyribonucleic acid from sturgeon gonads on three cases of moderate to severe oral mucositis in patients with head and neck cancer. three patients who had undergone rt and/or chemotherapy received an oral spray containing sodium salt{-}based natural deoxyribonucleic acid (pdrn) for grade 3 oral mucositis. treatment continued for one month after the end of rt. no patient reported any allergic reactions. rt and chemotherapy were not interrupted and opioid therapy was not given to any patient. pain was relieved about 2{-}3 days after starting treatment and oral mucositis was reduced to g2 within one week. outcomes in all 3 cases showed topical use of the sodium salt{-}based pdrn derived from sturgeon gonads was acceptable and safe when used topically for therapeutic and regenerative purposes.present results are encouraging and suggest a more in{-}depth study is warranted on its use in a larger patient cohort with rt{-}induced oral mucositis.\newline%
Answer: serious: 1 patientsex: 2 drugs: cisplatin reactions: candida infection, dehydration, pain, stomatitis, weight decreased\newline%
\newline%
Question: What adverse drug event was described in the following context?\newline%
Context: background piperacillin/tazobactam is a commonly used antibiotic for the empirical treatment of severe diabetic foot infections. one of the most feared complications of this drug is the development of pancytopenia. the aim of this study was to determine whether the use of piperacillin/tazobactam caused any hematological changes in patients admitted with severe diabetes{-}related foot infections from a specialist multidisciplinary foot clinic. specifically, looking at whether it caused anemia, leukopenia, neutropenia, or thrombocytopenia.   methods a 1{-}year retrospective analysis of patients admitted to a tertiary care center for treatment of diabetes{-}related foot infection using piperacillin/tazobactam. hematological indices, urea and electrolytes, and c{-}reactive protein (crp) were recorded pretreatment, during treatment, and posttreatment. hba1c, vitamin b12, folate, thyroid{-}stimulating hormone, and free thyroxin were also analyzed to exclude any potential confounders as a cause of pancytopenia.   results a total of 154 patients were admitted between 1 january 2016 and 31 december 2016 who received piperacillin/tazobactam for severe diabetes{-}related foot infection. on admission, white cell count and crp were raised and fell significantly within the first 48~h. other hematological factors did not change. five patients developed a mild pancytopenia, of which three were unexplained.   conclusions in this relatively small cohort, pancytopenia did not occur. as such, piperacillin/tazobactam appeared to have a low risk of adverse hematological outcomes and remains the treatment of choice for severe diabetes{-}related foot infections.\newline%
Answer: serious: 1 patientsex: 1 drugs: piperacillin sodium\textbackslash{}tazobactam sodium reactions: haemoglobin decreased, pancytopenia\newline%
\newline%
Question: What adverse drug event was described in the following context?\newline%
Context: \{\{full{-}text paper (as many tokens as possible)\}\} \newline%
Answer:

%% file: tables/prediction_comparison.tex
\begin{longtable}{@{}p{1.00\textwidth}@{}}
\toprule
\textbf{\OurDataset{} example} \tabularnewline
\midrule
\endhead
\bottomrule
\endfoot
\endlastfoot
(PMID: 28491911) TITLE: Navigating Long{-}Term Care. ABSTRACT: Americans over age 65 constitute a larger percentage of the population each year: from 14\% in 2010 (40 million elderly) to possibly 20\% in 2030 (70 million elderly). In 2015, an estimated 66 million people provided care to the ill, disabled, and elderly in the United States. In 2000, according to the Centers for Disease Control and Prevention (CDC), 15 million Americans used some form of long{-}term care: adult day care, home health, nursing home, or hospice. In all, 13\% of people over 85 years old, compared with 1\% of those ages 65 to 74, live in nursing homes in the United States. Transitions of care, among these various levels of care, are common: Nursing home to hospital transfer, one of the best{-}studied transitions, occurs in more than 25\% of nursing home residents per year. This article follows one patient through several levels of care. TEXT: Case: AB Mrs. AB is an 84{-}year{-}old Caucasian female with a history of hypertension, osteoporosis, type 2 diabetes, dyslipidemia, osteoarthritis, and persistent depression who presents to the office as a new patient with worsening ambulation: “I’m just not getting around well.” The patient lives in a small house above the family farm, on the side of a mountain. She describes her difficulty as an unsteadiness, and stiffness, in her knees and hips. She has moderate pain in her right hip and in her left knee, especially late in the day. On clinical examination, AB has reduced internal and external rotation of the hips, right side more affected than the left, with some pain to the maneuvers, and widened knees with some tenderness. A Mini{-}Mental Status Exam (MMSE) is consistent with mild cognitive impairment, with a score of 24. (Generally, scores of 27{-}30 are normal, 24{-}26 suggest mild cognitive impairment, 19{-}23 mild dementia, 10{-}18 moderate dementia, and <10 severe dementia.) She is taking 17 different medications, listed in the box below. AB has a son, Fred, who lives in the main farmhouse below her house, but AB does not get along with him well: He has a diagnosis of bipolar, and they argue frequently. Her other son, Rod, lives in Texas, and has recently been diagnosed with leukemia. Rod helps her with medical decisions—For example, he helped her pick her current Medicare part D plan. AB also has one surviving brother, 89 years old, but he is rather debilitated. He lives close by her house, but is unable to assist her; in fact, she assists him—sh... [Truncated] \\ \\ \begin{tabular}{lll} \hline & serious & patientsex \\ \hline target & 1 & 2 \\ flan-t5-large & 1 & 2 \\ gpt-4 & 1 & 2 \\ \hline & \multicolumn{2}{l}{drugs} \\ \hline target & \multicolumn{2}{p{13.2cm}}{alendronate sodium, amitriptyline, ascorbic acid, celecoxib, chromic chloride\textbackslash{}chromium, cinnamon, diltiazem, ginkgo, glucosamine, glyburide, hydroxyzine hydrochloride, metformin hydrochloride, niacin, paroxetine, pioglitazone, simvastatin, st. john\^{}s wort} \\ flan-t5-large & \multicolumn{2}{p{13.2cm}}{alendronate sodium, amitriptyline, celecoxib, glimepiride, hydroxyzine palmitate, niacin, paroxetine, simvastatin} \\ gpt-4 & \multicolumn{2}{p{13.2cm}}{diltiazem xr, simvastatin, amitriptyline, paroxetine, st. john's wort, celecoxib, metformin, alendronate, glyburide xr, pioglitazone, hydroxyzine palmoate, chromium, cinnamon, ginkgo, glucosamine, niacin, vitamin c} \\ \hline & \multicolumn{2}{l}{reactions} \\ \hline target & \multicolumn{2}{p{13.2cm}}{cognitive disorder, delirium, dementia alzheimer\^{}s type, drug interaction, fall, hip fracture, mobility decreased} \\ flan-t5-large & \multicolumn{2}{p{13.2cm}}{drug interaction, memory impairment} \\ gpt-4 & \multicolumn{2}{p{13.2cm}}{cognitive impairment, drug{-}drug interactions, polypharmacy} \\ \hline \end{tabular} \\ \\
 (PMID: 32695989) TITLE: The Efficacy of Albumin Dialysis in the Reversal of Refractory Vasoplegic Shock Due to Amlodipine Toxicity. ABSTRACT: Calcium channel blockers are highly protein{-}bound medications frequently used in the management of hypertension. Overdose results in severe hypotension and is the fourth most common cause of toxicity{-}related deaths in the United States. Management is mostly supportive, with currently no standard role for targeted drug removal. The protein{-}bound nature of these medications presents the option of utilizing albumin dialysis for their removal and for the reversal of associated shock. We present two cases of life{-}threatening intentional amlodipine overdoses successfully treated with albumin dialysis. Both patients experienced profound distributive shock in the setting of preserved cardiac contractility that was refractory to maximal vasoactive agent support. After initiation of albumin dialysis, the patients showed rapid hemodynamic improvement and were able to be weaned off vasopressor support. These cases demonstrate the safety and efficacy of albumin dialysis in the management of near{-}fatal calcium channel blocker overdoses related to amlodipine and offer an additional therapeutic option apart from conventional supportive care. Importantly, these cases were not associated with impaired cardiac contractility, thereby making venoarterial extracorporeal membrane oxygenation a less preferable option. Furthermore, this therapeutic benefit of albumin dialysis can potentially be extended to the management of toxicity related to other highly protein{-}bound drugs and toxins. TEXT: According to the National Poison Data System, calcium channel blocker (CCB) toxicity was the fourth highest cause of toxicity{-}related deaths in 2016, accounting for over 5\% of fatal exposures (1). Non{-}dihydropyridine CCB (e.g., verapamil, diltiazem) toxicity can cause negative inotropic and chronotropic effects, in particular, resulting in life{-}threatening cardiogenic shock. Typical therapies are supportive, aimed to temporize hemodynamic derangements until inherent elimination can occur. Venoarterial extracorporeal membrane oxygenation (VA{-}ECMO) is an additional therapeutic option in this context. In contrast, dihydropyridine CCB (e.g., amlodipine) toxicity is predominantly associated with systemic vasodilation and less cardiac depression, thereby resulting in distributive shock; in this setting, VA{-}ECMO has not traditionally been used given pre... [Truncated] \\ \\ \begin{tabular}{lll} \hline & serious & patientsex \\ \hline target & 1 & 1 \\ flan-t5-large & 1 & 2 \\ gpt-4 & 1 & 1 \\ \hline & \multicolumn{2}{l}{drugs} \\ \hline target & \multicolumn{2}{p{13.2cm}}{amlodipine besylate} \\ flan-t5-large & \multicolumn{2}{p{13.2cm}}{amlodipine besylate, lisinopril} \\ gpt-4 & \multicolumn{2}{p{13.2cm}}{amlodipine, lisinopril} \\ \hline & \multicolumn{2}{l}{reactions} \\ \hline target & \multicolumn{2}{p{13.2cm}}{intentional overdose, shock} \\ flan-t5-large & \multicolumn{2}{p{13.2cm}}{intentional overdose, metabolic acidosis, shock} \\ gpt-4 & \multicolumn{2}{p{13.2cm}}{refractory vasoplegic shock, hypotension, overdose} \\ \hline \end{tabular} \\ \\
 (PMID: 33363981) TITLE: Cyanide poisoning in inhalation injuries. ABSTRACT: Cyanide gas forms during the combustion of synthetic polymers and should be considered in patients presenting with inhalation injuries. A persistently high lactate following adequate resuscitation may be an indicator of cyanide exposure. As cyanide poisoning can be rapidly fatal, prompt recognition and treatment of this condition is vital. TEXT: A 78‐year‐old man was admitted to a National Burns Unit following a 22\% total body surface area flame burn and inhalation injury. This occurred following an explosion while lighting a gas fire in his outhouse. Despite adequate fluid resuscitation and good baseline renal function, a severe increased anion gap metabolic acidosis, with an associated elevated lactate (2.26~mmol/L), persisted. Cyanide poisoning was suspected, and hydroxocobalamin was administered. Following administration, his urine rapidly turned a characteristic red‐wine color (Figure~1). Cyanide is a mitochondrial toxin which preferentially binds ferric ions in cytochrome oxidase a3—inhibiting this final enzyme in the mitochondrial cytochrome complex. This causes oxidative phosphorylation to cease. Cells switch to anaerobic metabolism leading to the formation of lactic acid and a metabolic acidosis. 1 Hydroxocobalamin is a synthetic form of vitamin B12 which binds cyanide and forms the nontoxic cyanocobalamin. This is renally cleared, giving the urine a dark red color. Onset of chromaturia typically occurs within the first 2~hours following administration and can persist for up to 35~days. 2 Figure 1 Red‐wine colored urine as a result of hydroxocobalamin administration Cyanide gas forms during the combustion of synthetic polymers often found in building materials and furnishings. As cyanide gas can be rapidly fatal, a low threshold for treatment should exist in those suspected of having inhalation injuries. Within 7~hours of administration of hydroxocobalamin, the patient's acidosis had resolved and his lactate had significantly improved (1.49~mmol/L). As expected, his urine remained discolored for approximately three weeks. After a protracted hospital stay, the patient was discharged home well and has since returned to work in his family business. CONFLICT OF INTEREST None declared. AUTHOR CONTRIBUTIONS SK: drafted and reviewed the article. KC: reviewed the article. ETHICAL APPROVAL The regional Research Ethics Committee judged that this work was exempt from ethical review. ACKNOWLEDGMENT... [Truncated] \\ \\ \begin{tabular}{lll} \hline & serious & patientsex \\ \hline target & 2 & 1 \\ flan-t5-large & 1 & 1 \\ gpt-4 & 1 & 1 \\ \hline & \multicolumn{2}{l}{drugs} \\ \hline target & \multicolumn{2}{p{13.2cm}}{hydroxocobalamin} \\ flan-t5-large & \multicolumn{2}{p{13.2cm}}{hydroxocobalamin} \\ gpt-4 & \multicolumn{2}{p{13.2cm}}{hydroxocobalamin} \\ \hline & \multicolumn{2}{l}{reactions} \\ \hline target & \multicolumn{2}{p{13.2cm}}{chromaturia} \\ flan-t5-large & \multicolumn{2}{p{13.2cm}}{blood chromaturia, red urine} \\ gpt-4 & \multicolumn{2}{p{13.2cm}}{cyanide poisoning, inhalation injury, metabolic acidosis} \\ \hline \end{tabular} \\ \\
 (PMID: 32373453) TITLE: EBV{-}associated lymphoid interstitial pneumonia in IBD patient: Case report and literature review. ABSTRACT: Lymphoid interstitial pneumonia (LIP) is categorized as a rare form of interstitial lung disease. Most cases are associated with autoimmune disease. A 78{-}year{-}old male with Crohn's disease, presented with progressive dyspnea and dry cough for few weeks. The pathology of transbronchial lung biopsy was compatible with LIP and positive cells on EBER in situ hybridization. Blood EBV viral load was 85,715 copies/mL, compatible with EBV{-}associated LIP. All immunosuppressive agents were discontinued, but unfortunately the patient died due to hospital{-}acquired infections. In addition, we reviewed all reported cases of EBV{-}associated LIP in literature. To our knowledge, we report herein the first case of EBV{-}associated LIP in an IBD patient. We postulate that LIP was the consequence from EBV reactivation, probably due to immunosuppressive agents and/or IBD itself. The physician should aware of this disease when taking care of immunosuppressive patients who present with acute interstitial pneumonitis. TEXT: 1 Introduction Lymphoid interstitial pneumonia (LIP) is categorized as a rare form of interstitial lung disease according to the classification of American Thoracic Society/European Respiratory Society {[}1{]}. The definite diagnosis requires both imagings and pathology. Chest computed tomogram reveals the presence of ground glass attenuation, centrilobular and subpleural nodules, and thickening of bronchovascular bundles. The pathologic are characterized by the presence of dense polyclonal interstitial lymphocytic infiltrates with widening interlobular and alveolar septa {[}2,3{]}. Most cases are associated with autoimmune disease or lymphoproliferative disorder {[}4{]}. EBV, a double{-}stranded DNA virus, belongs to the Herpesviridae family {[}5{]}. EBV is able to cause latent infection, and reactivation occurs when infected individuals develop immunosuppressive state. Primary EBV infection causes infectious mononucleosis syndrome, and chronic infection/reactivation can cause lymphoma, and lymphoproliferative disorder including post transplant lymphoproliferative disease (LPD) {[}6{]}. In latent phase of infection, viral protein has the ability to transform mature B lymphocyte, resulting in uncontrolled its proliferation, as LPD {[}7{]}. Inflammatory bowel diseases (IBDs), including Crohn's disease and ulcerative colitis, have been re... [Truncated] \\ \\ \begin{tabular}{lll} \hline & serious & patientsex \\ \hline target & 1 & 1 \\ flan-t5-large & 1 & 1 \\ gpt-4 & 1 & 1 \\ \hline & \multicolumn{2}{l}{drugs} \\ \hline target & \multicolumn{2}{p{13.2cm}}{azathioprine, infliximab, mesalamine, prednisolone} \\ flan-t5-large & \multicolumn{2}{p{13.2cm}}{azathioprine, infliximab, mesalamine, prednisolone} \\ gpt-4 & \multicolumn{2}{p{13.2cm}}{azathioprine, ganciclovir, infliximab, mesalazine, prednisolone} \\ \hline & \multicolumn{2}{l}{reactions} \\ \hline target & \multicolumn{2}{p{13.2cm}}{epstein{-}barr virus infection reactivation, idiopathic interstitial pneumonia} \\ flan-t5-large & \multicolumn{2}{p{13.2cm}}{acute respiratory failure, interstitial lung disease} \\ gpt-4 & \multicolumn{2}{p{13.2cm}}{autoimmune hemolytic anemia, cytomegalovirus colitis, lymphoid interstitial pneumonia, respiratory failure} \\ \hline \end{tabular} \\ \\
 (PMID: 32493855) TITLE: Reversible Cancer Therapeutics{-}related Cardiac Dysfunction Complicating Intra{-}cardiac Thrombi. ABSTRACT: Epirubicin{-}based chemotherapy carries a risk of inducing heart failure, although the frequency is rare. Bevacizumab, an anti{-}vascular endothelial growth factor monoclonal antibody, has recently been widely used in patients with recurrent breast cancer as a first{-}line chemotherapeutic agent. Heart failure or arterial thromboembolism has been reported as a rare cardiovascular complication of bevacizumab. We herein report a breast cancer patient with reversible cancer therapeutics{-}related cardiac dysfunction associated with bevacizumab and epirubicin complicating intracardiac thrombi in the left atrium and left ventricle. This case underscores the importance of tailored medical planning according to the individual status in patients receiving anti{-}cancer therapies. TEXT: Introduction Anthracycline, including epirubicin{-}based chemotherapy, improves the survival of breast cancer patients but is associated with an increased risk of heart failure (1). In recent years, systemic therapy targeting vascular endothelial growth factor (VEGF) and its receptors has proven to be a successful strategy in patients with cancer. Bevacizumab is a widely used anti{-}VEGF monoclonal antibody targeting the VEGF ligand. Although it has been shown to improve clinical outcomes in several malignancies including advanced breast cancer (2), its use has been associated with many cardiovascular events (3{-}5). We herein report a breast cancer patient with reversible cancer therapeutics{-}related cardiac dysfunction associated with bevacizumab along with epirubicin complicated by intracardiac thrombi in the left atrium and left ventricle. Case Report A 65{-}year{-}old woman with a history of postoperative chemotherapy for right breast cancer was referred to our department due to congestive heart failure. The breast cancer had been graded as clinical stage IIa, triple{-}negative invasive ductal carcinoma {[}estrogen receptor 0\%, progressive receptor 0\%, and human epidermal growth factor receptor 2 (HER2) immunohistochemistry 0\%{]}, and the Ki{-}67{-}positive cell index was 98.6\%. She had received 4 courses of epirubicin (total dose: 327 mg/m2) and cyclophosphamide (total dose: 2,183 mg/m2) followed by paclitaxel (total dose: 727 mg/m2) and bevacizumab (total dose: 546 mg/m2). Nine months after the end of epirubicin administration and three months after the end... [Truncated] \\ \\ \begin{tabular}{lll} \hline & serious & patientsex \\ \hline target & 1 & 2 \\ flan-t5-large & 1 & 2 \\ gpt-4 & 1 & 2 \\ \hline & \multicolumn{2}{l}{drugs} \\ \hline target & \multicolumn{2}{p{13.2cm}}{bevacizumab, cyclophosphamide, epirubicin, paclitaxel} \\ flan-t5-large & \multicolumn{2}{p{13.2cm}}{bevacizumab, cyclophosphamide, epirubicin, paclitaxel} \\ gpt-4 & \multicolumn{2}{p{13.2cm}}{epirubicin, bevacizumab} \\ \hline & \multicolumn{2}{l}{reactions} \\ \hline target & \multicolumn{2}{p{13.2cm}}{bundle branch block right, cardiac failure, intracardiac thrombus, pleural effusion} \\ flan-t5-large & \multicolumn{2}{p{13.2cm}}{cardiac failure, cardiac thrombosis, cardiomegaly, bundle branch block right, dyspnoea exertional, left ventricular hypertrophy, left atrial thrombosis, left ventricular dysfunction, sinus tachycardia, sinus thrombosis, sinus thorax, ventricular hypertrophy} \\ gpt-4 & \multicolumn{2}{p{13.2cm}}{cancer therapeutics{-}related cardiac dysfunction, heart failure, intracardiac thrombi} \\ \hline \end{tabular} \\ \\
 (PMID: 31123688) TITLE: Ceftaroline{-}Associated Neutropenia: Case Series and Literature Review of Incidence, Risk Factors, and Outcomes. ABSTRACT: Ceftaroline is increasingly prescribed for "off{-}label" indications involving longer durations and higher doses. There have been postmarketing case reports of neutropenia among patients who have received extended durations of ceftaroline, but limited published data currently exist on its incidence and risk factors. We review a total of 37 published cases of ceftaroline{-}associated neutropenia including cases (n = 4) identified in our health care system. The median time from ceftaroline initiation to development of neutropenia (range) was 25 (8{-}125) days, with a median duration of neutropenia (range) of 4 (1{-}16) days. Agranulocytosis (absolute neutrophil count {[}ANC{]} nadir < 100 cells/mm3) developed in 49\% of cases (n = 18), and there was an ANC nadir of 0 in 27\% (n = 10). The overall incidence of neutropenia among cases receiving ceftaroline for $\geq$7{-}14 days (range) was 12\% (7\%{-}18\% per individual study), higher than for comparator antibiotics in the literature. Risk factors for ceftaroline{-}associated neutropenia varied among studies and remain poorly defined. TEXT: The development of novel antibiotics is important in addressing the growing rates of antibiotic resistance. For instance, Staphylococcus aureus remains a leading cause of bacteremia and endocarditis, with an increasing preponderance due to methicillin{-}resistant S. aureus (MRSA) strains {[}1, 2{]}. Given the limitations of the currently available antibiotics (eg, vancomycin) for treating MRSA infections, including drug intolerance, adverse events, and/or clinical failure {[}3, 4{]}, new antibiotics with anti{-}MRSA activity have been recently developed. Ceftaroline (Teflaro®) gained Food and Drug Administration (FDA) approval in 2010 and is the first licensed cephalosporin that includes coverage against MRSA. Studies leading to its approval include 2 clinical trials on community{-}acquired bacterial pneumonia (CABP; FOCUS 1 and FOCUS 2) {[}5, 6{]} and 2 additional studies on acute bacterial skin and skin structure infections (ABSSSIs; CANVAS 1 and CANVAS 2) {[}7, 8{]}. These 4 studies evaluated a total of 1307 subjects, with the most common adverse events among those receiving ceftaroline being diarrhea, nausea, and rash; no patient developed neutropenia. All studies utilized a ceftaroline dosage of 600 mg intravenously (IV) every 12 hours for dura... [Truncated] \\ \\ \begin{tabular}{lll} \hline & serious & patientsex \\ \hline target & 1 & 1 \\ flan-t5-large & 1 & 1 \\ gpt-4 & 1 & 1 \\ \hline & \multicolumn{2}{l}{drugs} \\ \hline target & \multicolumn{2}{p{13.2cm}}{ceftaroline fosamil, daptomycin, famotidine, linezolid, vancomycin} \\ flan-t5-large & \multicolumn{2}{p{13.2cm}}{ceftaroline hydrochloride} \\ gpt-4 & \multicolumn{2}{p{13.2cm}}{ceftaroline} \\ \hline & \multicolumn{2}{l}{reactions} \\ \hline target & \multicolumn{2}{p{13.2cm}}{eosinophilia, neutropenia, pancytopenia} \\ flan-t5-large & \multicolumn{2}{p{13.2cm}}{neutropenia} \\ gpt-4 & \multicolumn{2}{p{13.2cm}}{neutropenia} \\ \hline \end{tabular} \\ \\
 (PMID: 24995045) TITLE: A case of bilateral human herpes virus 6 panuveitis with genomic viral DNA integration. ABSTRACT: BACKGROUND We report a rare case of bilateral panuveitis from human herpes virus 6 (HHV{-}6) with genomic viral DNA integration in an immunocompromised man. RESULTS A 59{-}year{-}old man with history of multiple myeloma presented with altered mental status, bilateral eye redness, and blurry vision. Examination revealed bilateral diffuse keratic precipitates, 4+ anterior chamber cell, hypopyon, vitritis, and intraretinal hemorrhages. Intraocular fluid testing by polymerase chain reaction (PCR) was positive for HHV{-}6. The patient was successfully treated with intravitreal foscarnet and intravenous ganciclovir and foscarnet. Despite clinical improvement, his serum HHV{-}6 levels remained high, and it was concluded that he had HHV{-}6 chromosomal integration. CONCLUSIONS HHV{-}6 should be considered in the differential for infectious uveitis in immunocompromised hosts who may otherwise have a negative work{-}up. HHV{-}6 DNA integration may lead to difficulties in disease diagnosis and determining disease resolution. TEXT: Findings Human herpes virus{-}6 (HHV{-}6) is a ubiquitous virus that infects most children by the age of three years. While the seroprevalence in the adult population approaches 95\%, and HHV{-}6 reactivations are known to be common after organ transplantation, clinical disease is rare after the primary infection {[}1{]}. Although HHV{-}6 is closely related to cytomegalovirus (CMV), ocular disease due to HHV{-}6 has been described in very few patients {[}2{-}8{]}. We report the case of an immunocompromised man who presented with encephalitis and severe bilateral panuveitis as a result of HHV{-}6 reactivation. Integration of the viral genome into the host DNA, a unique characteristic of HHV{-}6, complicated the clinical management of our patient. Case report A 59{-}year{-}old man with a history of multiple myeloma status post allogeneic stem cell transplant was admitted to our hospital with fevers and a soft tissue infection. On the fourth day of hospitalization, he developed a headache, somnolence, bilateral eye redness, and blurred vision. On presentation, his best{-}corrected visual acuity was 20/100 in the right eye and unobtainable in the left eye due to his altered mental status. The pupils were equal bilaterally with a brisk direct response and no relative afferent pupillary defect. His intraocular pressure was 5~mmHg bilaterally... [Truncated] \\ \\ \begin{tabular}{lll} \hline & serious & patientsex \\ \hline target & 1 & 1 \\ flan-t5-large & 1 & 1 \\ gpt-4 & 1 & 1 \\ \hline & \multicolumn{2}{l}{drugs} \\ \hline target & \multicolumn{2}{p{13.2cm}}{ceftazidime, foscarnet sodium, ganciclovir, vancomycin} \\ flan-t5-large & \multicolumn{2}{p{13.2cm}}{ceftazidime, foscarnet, ganciclovir, vancomycin} \\ gpt-4 & \multicolumn{2}{p{13.2cm}}{foscarnet, ganciclovir} \\ \hline & \multicolumn{2}{l}{reactions} \\ \hline target & \multicolumn{2}{p{13.2cm}}{hypersensitivity vasculitis, off label use, renal impairment} \\ flan-t5-large & \multicolumn{2}{p{13.2cm}}{leukocytoclastic vasculitis, off label use, renal impairment} \\ gpt-4 & \multicolumn{2}{p{13.2cm}}{encephalitis, leukocytoclastic vasculitis, panuveitis, renal impairment} \\ \hline \end{tabular} \\ \\
 (PMID: 25888368) TITLE: Chromosomal rearrangement involving 11q23 locus in chronic myelogenous leukemia: a rare phenomenon frequently associated with disease progression and poor prognosis. ABSTRACT: BACKGROUND Progression of chronic myelogenous leukemia (CML) is frequently accompanied by cytogenetic evolution, commonly unbalanced chromosomal changes, such as an extra copy of Philadelphia chromosome (Ph), +8, and i(17)(q10). Balanced chromosomal translocations typically found in de novo acute myeloid leukemia occur occasionally in CML, such as inv(3)/t(3;3), t(8;21), t(15;17), and inv(16). Translocations involving the 11q23, a relatively common genetic abnormality in acute leukemia, have been seldom reported in CML. In this study, we explored the prevalence and prognostic role of 11q23 in CML. METHODS We searched our pathology archives for CML cases diagnosed in our institution from 1998 to present. Cases with 11q23 rearrangements were retrieved. The corresponding clinicopathological data were reviewed. RESULTS A total of 2,012 cases of CML with available karyotypes were identified. Ten (0.5\%) CML cases had 11q23 rearrangement in Ph{-}positive cells, including 4 cases of t(9;11), 2 cases of t(11;19), and 1 case each of t(2;11), t(4;11), t(6;11), and t(4;9;11). Eight cases (80\%) had other concurrent chromosomal abnormalities. There were 6 men and 4 women with a median age of 50 years (range, 21{-}70 years) at time of initial diagnosis of CML. 11q23 rearrangement occurred after a median period of 12.5 months (range, 0{-}172 months): 1 patient in chronic phase, 2 in accelerated phase, and 7 in blast phase. Eight of ten patients died after a median follow{-}up of 16.5 months (range, 8{-}186 months) following the initial diagnosis of CML, and a median of 6.7 months (range, 0.8{-}16.6 months) after the emergence of 11q23 rearrangement. The remaining two patients had complete remission at the last follow{-}up, 50.2 and 6.9 months, respectively. In addition, we also identified a case with 11q23/t(11;17) in Ph{-}negative cells in a patient with a history of CML. MLL involvement was tested by fluorescence in situ hybridization in 10 cases, and 7 cases (70\%) were positive. CONCLUSIONS In summary, chromosomal rearrangements involving 11q23 are rare in CML, frequently occurring in blast phase, and are often associated with other cytogenetic abnormalities. These patients had a low response rate to tyrosine kinase inhibitors and a poor prognosis. TEXT: Background BCR{-}ABL1 derived... [Truncated] \\ \\ \begin{tabular}{lll} \hline & serious & patientsex \\ \hline target & 1 & 1 \\ flan-t5-large & 1 & 1 \\ gpt-4 & 1 & 1 \\ \hline & \multicolumn{2}{l}{drugs} \\ \hline target & \multicolumn{2}{p{13.2cm}}{bosutinib, dasatinib, imatinib, nilotinib} \\ flan-t5-large & \multicolumn{2}{p{13.2cm}}{hydroxyurea, imatinib} \\ gpt-4 & \multicolumn{2}{p{13.2cm}}{imatinib, dasatinib, nilotinib} \\ \hline & \multicolumn{2}{l}{reactions} \\ \hline target & \multicolumn{2}{p{13.2cm}}{blast cell count increased, chronic myeloid leukaemia transformation, drug ineffective, thrombocytopenia} \\ flan-t5-large & \multicolumn{2}{p{13.2cm}}{blast stage leukaemia} \\ gpt-4 & \multicolumn{2}{p{13.2cm}}{clonal evolution, disease progression, poor prognosis} \\ \hline \end{tabular} \\ \\
 (PMID: 29170802) TITLE: Pharmacokinetics and safety of panitumumab in a patient with chronic kidney disease. ABSTRACT: Data on panitumumab dosing in cancer patients with renal insufficiency are lacking. Here, we report a 63{-}year{-}old metastatic colorectal cancer patient with chronic kidney injury with a glomerular filtration rate of approximately 11~mL/min. Pharmacokinetic parameters, including dose{-}normalized area under the curve, clearance and elimination half{-}life (T 1/2) after the 11th and 12th infusions were estimated using trapezoidal non{-}compartmental methods. Data were compared to previous reported pharmacokinetic data from studies in patients with normal renal function. The results show that the pharmacokinetic data in this patient with kidney failure are comparable to those in patients with adequate renal function. Moreover the treatment was well tolerated in this patient. This study suggests that panitumumab can be safely used in cancer patients with renal impairment without dose adjustment. TEXT: Introduction Panitumumab is a fully humane monoclonal antibody targeting the epidermal growth factor receptor (EGFR) and is registered for the treatment of RAS wild{-}type metastatic colorectal cancer, either alone or combined with chemotherapy. As previously discussed elsewhere, clearance of panitumumab mainly occurs by an EGFR sink. In case of saturation of all receptors, panitumumab will be cleared by immunologic mechanisms, such as complement{-}dependent cytotoxicity (CDC), antibody dependent cell{-}mediated cytotoxicity and apoptosis {[}1{]}. Therefore, theoretically renal insufficiency is not likely to influence the pharmacokinetics of panitumumab. The study of councilman et al. showed that nephrotic syndrome was associated with increased rituximab clearance, and therefore, decreased half{-}life. An possible explanation for the observed effect is loss of monoclonal antibody in the urine and not altered clearance {[}2{]}. The most recent summary of product characteristics (SmPc) of panitumumab states that a population pharmacokinetic analysis (among race, age, gender, hepatic function, concomitant chemotherapy and EGFR membrane{-}staining intensity in tumor cells) renal function does not influence the pharmacokinetics of panitumumab, however, it is not tested in patients. The only available clinical information concerns a case report showing safety and efficacy of panitumumab (combined with oxaliplatin, folic acid and 5{-}FU) in a hemodialysis patient ... [Truncated] \\ \\ \begin{tabular}{lll} \hline & serious & patientsex \\ \hline target & 1 & 1 \\ flan-t5-large & 1 & 1 \\ gpt-4 & 1 & 1 \\ \hline & \multicolumn{2}{l}{drugs} \\ \hline target & \multicolumn{2}{p{13.2cm}}{fluorouracil, folic acid, oxaliplatin} \\ flan-t5-large & \multicolumn{2}{p{13.2cm}}{fluorouracil, leucovorin, oxaliplatin, panitumumab} \\ gpt-4 & \multicolumn{2}{p{13.2cm}}{panitumumab} \\ \hline & \multicolumn{2}{l}{reactions} \\ \hline target & \multicolumn{2}{p{13.2cm}}{product use in unapproved indication, renal impairment} \\ flan-t5-large & \multicolumn{2}{p{13.2cm}}{electrolyte imbalance, skin toxicity} \\ gpt-4 & \multicolumn{2}{p{13.2cm}}{skin toxicity} \\ \hline \end{tabular} \\ \\
 (PMID: 24860718) TITLE: Cardiac safety results from a phase II, open{-}label, multicenter, pilot study of two docetaxel{-}based regimens plus bevacizumab for the adjuvant treatment of subjects with node{-}positive or high{-}risk node{-}negative breast cancer. ABSTRACT: OBJECTIVE Adding antiangiogenic therapy to standard chemotherapy has improved response rates and progression{-}free survival in metastatic breast cancer (BC) patients. This phase II study evaluated cardiac safety of bevacizumab with/without trastuzumab with two docetaxel{-}based regimens in early BC. METHODS 127 women with non{-}metastatic node{-}positive or high{-}risk node{-}negative BC were enrolled. Women with human epidermal growth factor receptor 2 (HER2){-}negative BC (n = 93) received docetaxel/doxorubicin/cyclophosphamide (TAC) + bevacizumab, while women with HER2{-}positive disease (n = 34) received docetaxel/carboplatin/trastuzumab (TCH) + bevacizumab, every 3~weeks for six cycles. Maintenance therapy with bevacizumab alone or bevacizumab plus trastuzumab, respectively, was given every 3~weeks for 52~weeks. The primary objective was to evaluate cardiac safety, as measured by the incidence of $\geq$ grade 3 clinical congestive heart failure (CHF); the secondary objective was assessment of safety and toxicity. RESULTS At least one cardiac adverse event (AE; CHF, cardiomyopathy, or left ventricular dysfunction) was reported in 26.1\% of TAC (n = 92) and 17.6\% of TCH subjects (n = 34); there were no cardiac deaths. $\geq$ Grade 3 clinical CHF was observed in 4.3\% in the TAC plus bevacizumab stratum and 0\% in the TCH plus bevacizumab stratum. A $\geq$ grade 3 treatment{-}emergent AE (any kind) related to study treatment was observed in 59.8\% in the TAC with bevacizumab and 52.9\% in the TCH plus bevacizumab stratum. CONCLUSIONS Adding bevacizumab to a docetaxel{-}based regimen with trastuzumab did not appear to increase cardiotoxicity. BACKGROUND ClinicalTrials.gov Identifier: NCT00446030, registered March 8, 2007. TEXT: Introduction Breast cancer mortality has declined over the past 2 decades; however, it still remains the most common type of cancer in women, accounting for an estimated 29\% of all new cases (Siegel et al. 2014). The 5{-}year survival rate for women with breast cancer is 99\% for those with localized disease and 84\% for regional disease, and only 24\% in patients with distant disease (Siegel et al. 2014). Several studies in human epidermal growth factor receptor 2 (HER2){-}normal metastatic... [Truncated] \\ \\ \begin{tabular}{lll} \hline & serious & patientsex \\ \hline target & 1 & 2 \\ flan-t5-large & 1 & 2 \\ gpt-4 & 1 & 2 \\ \hline & \multicolumn{2}{l}{drugs} \\ \hline target & \multicolumn{2}{p{13.2cm}}{bevacizumab, cyclophosphamide, docetaxel, doxorubicin hydrochloride} \\ flan-t5-large & \multicolumn{2}{p{13.2cm}}{bevacizumab, carboplatin, docetaxel, doxorubicin, trastuzumab} \\ gpt-4 & \multicolumn{2}{p{13.2cm}}{bevacizumab, trastuzumab, docetaxel, doxorubicin, cyclophosphamide, carboplatin} \\ \hline & \multicolumn{2}{l}{reactions} \\ \hline target & \multicolumn{2}{p{13.2cm}}{clostridial infection} \\ flan-t5-large & \multicolumn{2}{p{13.2cm}}{cardiac failure congestive} \\ gpt-4 & \multicolumn{2}{p{13.2cm}}{congestive heart failure, cardiomyopathy, left ventricular dysfunction} \\ \hline \end{tabular} \\\end{longtable}

%% file: figures/drug-classification.tex
\begin{figure}[t!]
\centering
\includegraphics[height=11cm]{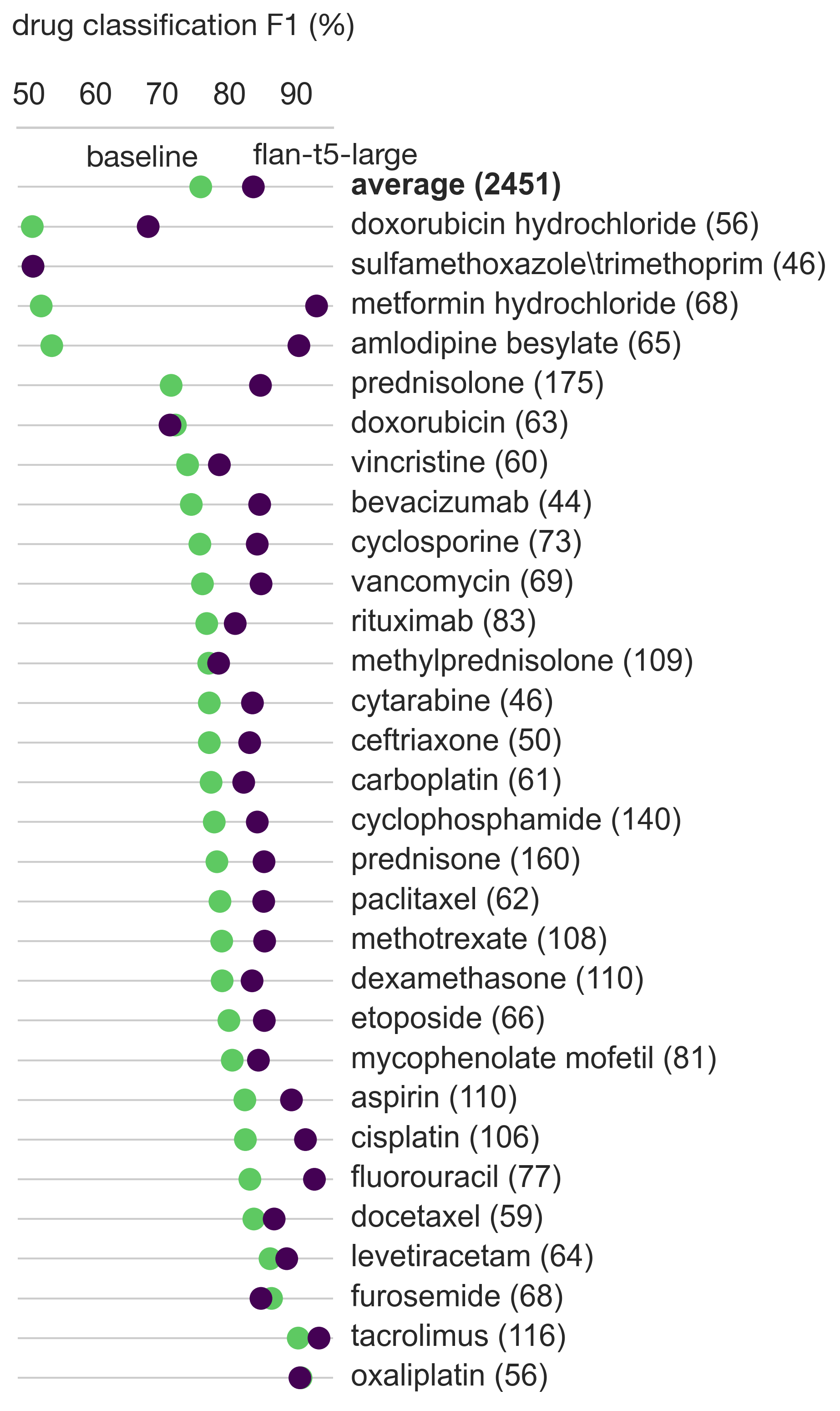}
\caption{Reaction classification performance across the 30 most frequent drugs in the \OurDataset{} \texttt{validation} set. Baseline performance in lighter color, \texttt{FLAN-T5} in darker color. Support in parentheses. Average performance in bold. Drugs are sorted by baseline performance.} \label{fig:drug-classification}
\end{figure}